\documentclass[10pt,twocolumn,letterpaper]{article}

\usepackage{cvpr}              

\usepackage[percent]{overpic}

\usepackage[dvipsnames]{xcolor}
\usepackage{bbm}

\newcommand{\R}{\mathbb{R}}

\newcommand{\boldstart}[1]{\vspace{0.06in}\noindent{\bf #1}}

\newcommand{\lightdir}{\hat{\boldsymbol{\omega}}_i}
\newcommand{\viewdir}{\hat{\boldsymbol{\omega}}_o}
\newcommand{\halfvec}{\hat{\boldsymbol{\omega}}_h}
\newcommand{\bomega}{\hat{\boldsymbol{\omega}}}
\newcommand{\brho}{\boldsymbol{\rho}}
\newcommand{\fzero}{\kappa}
\newcommand{\bu}{\mathbf{u}}

\newcommand{\bx}{\mathbf{x}}
\newcommand{\uvn}{\hat{\mathbf{n}}}
\newcommand{\params}{\boldsymbol{\phi}}
\newcommand{\brdfparams}{\boldsymbol{\alpha}}
\newcommand{\radius}{r}

\newcommand{\eqdef}{\overset{\Delta}{=}}

\definecolor{cvprblue}{rgb}{0.21,0.49,0.74}
\usepackage[pagebackref,breaklinks,colorlinks,citecolor=cvprblue]{hyperref}

\def\paperID{2258} 
\def\confName{CVPR}
\def\confYear{2024}

\begin{document}

\title{Eclipse: Disambiguating Illumination and Materials using Unintended Shadows} 

\author{
Dor Verbin$^{1}$
\quad
Ben Mildenhall$^1$
\quad
Peter Hedman$^1$
\\
Jonathan T. Barron$^1$
\quad
Todd Zickler$^{1, 2}$
\quad
Pratul P. Srinivasan$^1$
\\
\vspace{2mm}
{$^1$Google Research \quad $^2$Harvard University}
}

\twocolumn[{
\renewcommand\twocolumn[1][]{#1}
\maketitle
\vspace{-0.3in}
\begin{center}
    \includegraphics[width=\linewidth]{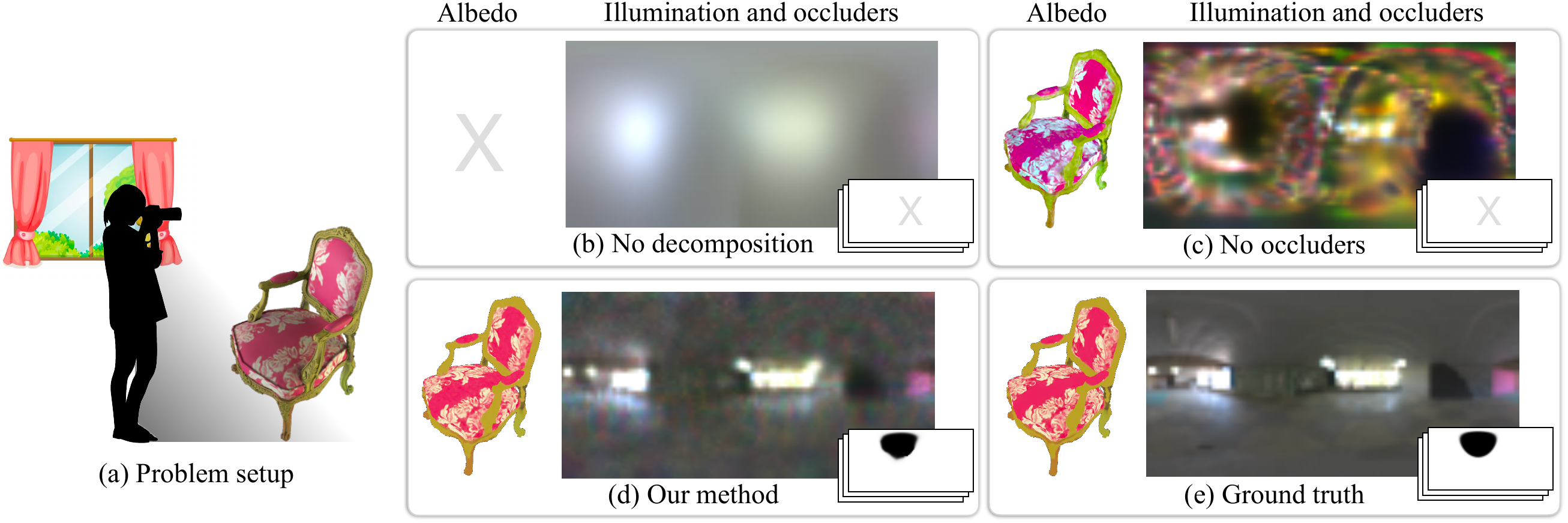}
\vspace{-0.25in}
\captionof{figure}{
(a) We exploit unintended shadows cast by camera operators (or other unobserved moving occluders) to recover high-fidelity environment lighting and object materials from a set of images. Without modeling such unobserved occluders, prior methods can only: (b) recover convolved lighting without explicit material decomposition~\cite{park2020chips}; or (c) exploit lighting occlusions that occur internally among the observed objects~\cite{swedish2021objects}. (d) We show that additionally modeling and recovering external, unobserved occluders enables lighting and material reconstructions that are closer to ground truth (e).
}\label{fig:teaser}
\end{center}
}]


\begin{abstract}

\vspace{-0.1in}

Decomposing an object's appearance into representations of its materials and the surrounding illumination is difficult, even when the object's 3D shape is known beforehand. This problem is especially challenging for diffuse objects: it is ill-conditioned because diffuse materials severely blur incoming light, and it is ill-posed because diffuse materials under high-frequency lighting can be indistinguishable from shiny materials under low-frequency lighting. We show that it is possible to recover precise materials and illumination---even from diffuse objects---by exploiting unintended shadows, like the ones cast onto an object by the photographer who moves around it. These shadows are a nuisance in most previous inverse rendering pipelines, but here we exploit them as signals that improve conditioning and help resolve material-lighting ambiguities. We present a method based on differentiable Monte Carlo ray tracing that uses images of an object to jointly recover its spatially-varying materials, the surrounding illumination environment, and the shapes of the unseen light occluders who inadvertently cast shadows upon it.
\end{abstract}

\section{Introduction}

In this work, we show that the long-standing inverse rendering problem of recovering an object's material properties and the surrounding illumination environment from a set of images can greatly benefit from considering the effects of unobserved moving light occluders, such as the camera operator, which partially block incoming light and inadvertently cast shadows onto the object being imaged. 

Joint recovery of materials and lighting is a challenging task because the BRDF (bidirectional reflectance distribution function, which represents how a material at any point on an object's surface maps from incoming to outgoing light) acts as a directional filter on incoming light. BRDFs for specular (shiny) materials act as all-pass directional filters, while BRDFs for diffuse materials act as low-pass directional filters. As a result, inverse rendering is fundamentally ambiguous since a shiny object illuminated by blurry lighting can be indistinguishable from a diffuse object illuminated by sharp lighting (Figure~\ref{fig:ambiguity}). Furthermore, since diffuse materials act as low-pass filters and strongly blur incoming light, even the simpler problem of recovering lighting from images of a known diffuse material can be severely ill-conditioned, precluding the recovery of high-frequency illumination~\cite{ramamoorthi2001spfw}.

We observe that both of these issues can be ameliorated by exploiting an effect that naturally exists whenever an object is imaged from a sequence of viewpoints under static environment lighting. Namely, between one image and the next, the positions of the camera and its operator(s) must change, and in doing so they occlude different portions of the surrounding environment, casting distinct sets of shadows onto the object being imaged. This effect is very noticeable under strongly-directional lighting or for shiny objects because the cast shadows are sharp or the specular highlights are misshapen and hidden. But even when the effect is barely perceptible, such as for diffuse objects under well-distributed lighting, there exists a subtle signal that one can take advantage of.

In this paper, we make first steps toward a practical algorithm that exploits this cue by using gradient descent with differentiable Monte Carlo ray tracing to jointly recover explicit representations of (i) spatially-varying reflectance, (ii) environment illumination, and (iii) the per-image shapes of any unseen, light-blocking occluders. We evaluate our algorithm using a challenging variety of simulated scenes, including scenes with diffuse-only materials and with object geometry that is not known beforehand. We also apply our method to an existing off-the-shelf dataset designed for inverse rendering, where the (unseen) camera rig moving to capture the scene acts as an accidental occluder. Our results demonstrate that external shadowing effects provide a strong and useful cue, even when they are subtle, and even without relying on strong domain-specific priors for the materials, illumination, or external occluder shapes. This suggests that incorporating unintended shadows into inverse rendering pipelines for real, captured data is valuable for improving the quality of material and lighting assets.

\section{Related Work}

We build on recent developments in differentiable ray tracing and on a long history of inverse rendering work, including the joint estimation of reflectance and lighting, and the estimation of lighting alone. We are also inspired by a separate line of work on passive non-line-of-sight imaging, which exploits similar light-blocking effects. 

\boldstart{Physics-based differentiable ray tracing.}
Modern differentiable ray tracers~\cite{li2018redner,mitsuba2} enable the computation of gradients of rendered images with respect to scene parameters (\ie, geometry, materials, and lighting) by differentiating through light transport simulation. Recent works have focused on improving efficiency and performance~\cite{Jakob2022DrJit,NimierDavid2020Radiative} and on accurately differentiating through visibility discontinuities---such as those caused by shadow and occlusion boundaries---with respect to shape and lighting~\cite{bangaru2022NeuralSDFReparam,Vicini2022sdf}. Our implementation of differentiable ray tracing leverages insights from these works as well as from Zeltner~\etal~\cite{zeltner2021MonteCarlo}, who provide design intuition for Monte Carlo differentiable ray tracers.

\boldstart{Inverse rendering of materials and lighting.}
Decomposing an object's appearance into representations of material and lighting is a long-standing problem in computer vision and graphics. In their foundational work, Ramamoorthi and Hanrahan~\cite{ramamoorthi2001spfw}
developed a signal processing approach by describing the outgoing light at a surface point as the spherical convolution of the BRDF and the incoming lighting. This formulation elucidates why inverse rendering is ill-posed and often ill-conditioned: it is ill-posed because there are multiple illumination-material pairs that convolve into the same image, and it is ill-conditioned because diffuse BRDFs act as a low-pass filters on lighting, which causes the estimates of medium- and high-frequency illumination to be very sensitive to noise.

Because of this, most approaches to inverse rendering rely on strong priors on materials and lighting. Single-view approaches have used hand-designed priors~\cite{barron2015sirfs} or priors in the form of neural networks trained with supervision from large datasets of material and lighting labels~\cite{li2020inverseindoor,li2018learning}. More relevant to us are multi-view approaches, which typically either assume known lighting~\cite{bi2020nrf,bi2020drv,srinivasan2021nerv} or rely on strong priors, such as assuming a single, highly-specular BRDF for the entire object~\cite{zhang2021physg} or lighting from a prior distribution that was pre-trained on a dataset of environment maps~\cite{boss2021nerd,zhang2021nerfactor}. The most closely related work is that for which lighting information is not provided as input~\cite{kuang2022neroic,munkberg2021nvdiffrec}.

For the sake of generality, we do not use such strong priors in our experiments because by avoiding them we can more directly measure the benefits that are gained by modeling unintended shadows as an additional cue. Because our model uses a generic gradient-descent framework, stronger application-specific priors can be added to it to improve performance on a specific domain of interest.

\boldstart{Estimating environment lighting.}
Another thread of inverse rendering research focuses on the task of recovering high-fidelity lighting environments, typically for augmented reality applications where virtual objects are rendered into photographs with consistent reflections and shadows. Seminal work by Debevec~\cite{debevec1998rendering} used images of a chrome sphere to measure environment lighting directly, and subsequent works have demonstrated that plausible environment lighting can be estimated without chrome spheres, using images of indoor~\cite{gardner2017indoorsingle,song2019neuralillumination} and outdoor scenes~\cite{holdgeoffroy2017deep,lalonde2012outdoor,legendre2019deeplight}, or images of a specific class of objects like faces~\cite{nishino2004eyes,legendre2020facelight}. These subsequent techniques use strong priors in the form of deep neural network weights that are trained with supervision to map from images to lighting, and they do not enforce physical rendering consistency between the recovered lighting and the observations. In contrast, we avoid strong priors and we explicitly enforce consistency.

More related to our approach is the work of Park~\etal~\cite{park2020chips}, which recovers physically-consistent environment maps from RGBD videos of shiny objects. However, their algorithm can only recover the convolution of the environment map with the BRDF of the observed object (\eg, Figure~\ref{fig:teaser}(b)), which precludes the recovery of high illumination frequencies unless the object is highly specular.

Also related is the work of Swedish~\etal~\cite{swedish2021objects}, which places a known, diffuse object on a ground plane (a scenario first studied by Sato~\etal~\cite{sato2003}) and uses its cast shadows to recover high-quality lighting. Their formulation is linear and it improves lighting estimates by leveraging self-shadowing among diffuse objects that have known geometry and albedo (\eg, Figure~\ref{fig:teaser}(c)). We generalize this by replacing the linear formulation with differentiable Monte Carlo ray tracing, which allows exploiting the additional shadowing effects caused by moving external occluders and leads to substantially improved results (\eg, Figure~\ref{fig:teaser}(d)). Our formulation also handles objects with more general spatially-varying BRDFs that are not known beforehand.

\boldstart{Passive non-line-of-sight imaging with occlusions.}
Passive non-line-of-sight techniques can also be seen as recovering environment illumination: They observe a reflective surface (which is typically diffuse and planar) and recover the appearance of a ``hidden scene'' from these reflections. Similar to us, prior works in this area have observed that the presence of an occluder between the observed surface and the hidden scene/environment aids recovery by introducing sharp angular variations into the rendering integral. This insight was first leveraged in settings with simple occluder-shapes like pinholes and pinspecks~\cite{torralba2012accidental} or corners joining walls ~\cite{bouman2017corners}. Subsequent work by Baradad~\etal~\cite{baradad2018using} generalizes this by considering light-occlusion effects caused by an arbitrary but known 3D shape (a task that is closely related to Swedish~\etal above). Yedidia~\etal~\cite{yedidia2019using} additionally recover the unobserved occluder's shape assuming it is a planar mask parallel to the observed planar surface. 

We generalize these prior works by using differentiable Monte Carlo rendering to replace their deconvolution-based algorithms, which are tailored to the specific case where the observed reflector is planar and diffuse. This allows using reflective objects that have arbitrary shapes, and arbitrary spatially-varying BRDFs which are not known beforehand. It also allows recovering the shapes of unobserved occluders that are arbitrary and time-varying.

\section{Motivation and Problem Setup} \label{section:setup}

\begin{figure}[t]
    \centering
    \includegraphics[width=\linewidth]{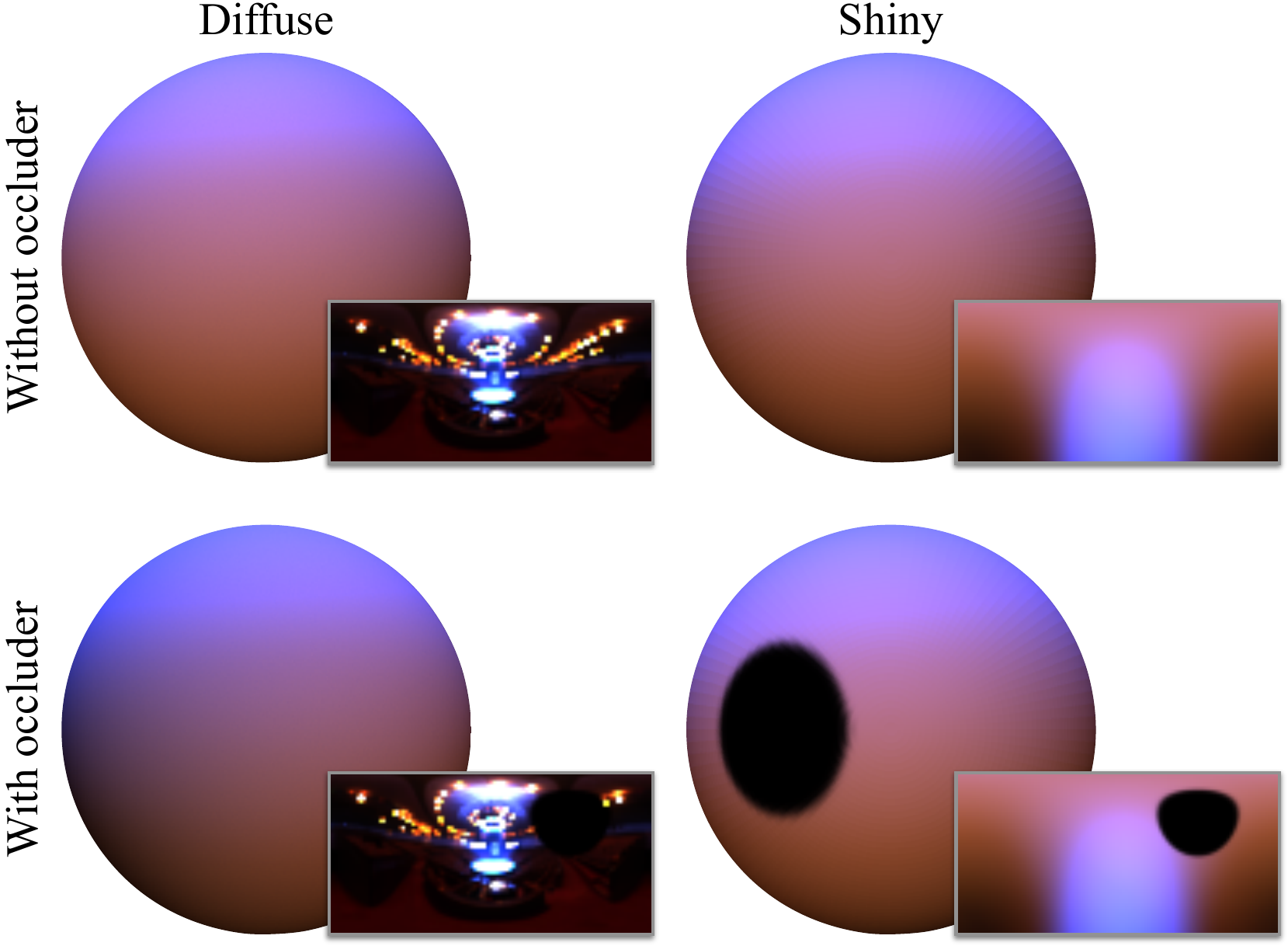}
    \caption{Decomposing appearance into lighting and materials can be inherently ambiguous. Top row: a diffuse sphere lit by high-frequency lighting (inset) is visually indistinguishable from a shiny, mirror-like sphere lit by low-frequency lighting. Bottom row: A second image captured in the presence of a dark external occluder resolves the ambiguity. The occluder's effect is clearly visible on the shiny sphere while on the diffuse sphere it produces shadows that are soft and very subtle.}
    \label{fig:ambiguity}
\end{figure}

A well-known ambiguity in computer vision and graphics occurs when decomposing an image into its component lighting and materials. The top of Figure~\ref{fig:ambiguity} recreates a common depiction of this ambiguity, where an image of a known shape (a sphere) is explained equally well by a diffuse material in a complicated lighting environment or a shiny, mirror-like material under low-frequency lighting. Now, imagine we capture a second image from the same viewpoint, after some external object enters the scene. This second object remains beyond the field of view and so is not directly observed, but it acts as an external occluder that prevents some of the environment's light from reaching the sphere. This second image, shown at the bottom of the figure for each case, clearly reveals which of the two material-lighting explanations is the correct one.

We aim to make use of this unintended shadowing that naturally occurs whenever a camera and its operators move around an object while capturing images from different viewpoints. They affect each image by blocking different portions of the surrounding light, and this provides a helpful signal. We will show that this signal is helpful even when the occluder shapes and locations are quite arbitrary and not known beforehand, and when their shadowing effects are very subtle, like between the top and bottom images on the left of Figure~\ref{fig:ambiguity}.

\begin{figure}[t]
    \centering
    \includegraphics[width=\linewidth]{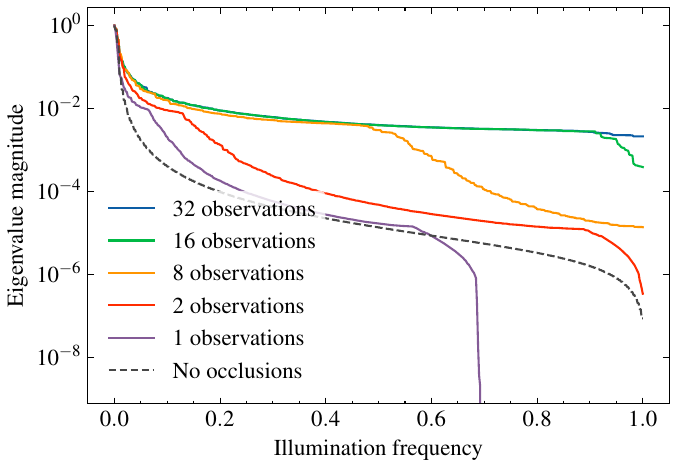}
    \vspace{-0.2in}    
    \caption{Here we plot eigenvalue magnitudes as a function of illumination angular frequency for a one-dimensional circular Lambertian object with known albedo, under the simplifying assumption of known occluder locations and shapes. For a single input image without occlusion, almost all lighting frequencies are strongly attenuated. In the single-occluder case (similar to the scenario considered by Baradad \etal~\cite{baradad2018using}), some low frequencies become recoverable but intermediate and high frequencies do not. Adding more occluders enables the recovery of more lighting information, especially at the highest frequencies.
    }
    \label{fig:eigenvalues}
\end{figure}

In addition to helping resolve material-lighting ambiguities, shadowing from moving external occluders also improves the conditioning of our inverse rendering problem. This is analyzed in Figure~\ref{fig:eigenvalues}, which considers the simplified hypothetical 1D case of recovering an environment illumination from images of a diffuse disk-shaped object, with known occluder positions. Convolution with a diffuse BRDF acts as a low-pass filter of the incoming light, which means that the linear inverse problem's eigenvalues are vanishingly low for higher frequencies. Moving external occluders introduce a sharp angular variation in the per-image illumination, which better-conditions these high frequency components, and allows their recovery. As shown in Figure~\ref{fig:eigenvalues}, in this simplified scenario a single image containing an occluder is better than a single image without any occluders, but using observations of more occluders makes the problem even better-conditioned. We demonstrate this effect in practice in the supplement.

\begin{figure*}[t]
    \centering
    \begin{tabular}{@{}c@{\,\,}c@{\,\,}c@{\,\,}@{\,}c@{\,\,}c@{}}
    \includegraphics[width=0.14\linewidth]{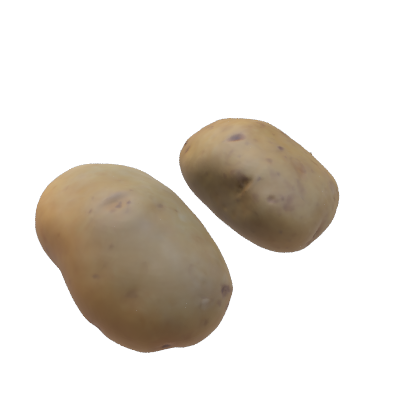} & 
    \includegraphics[width=0.14\linewidth]{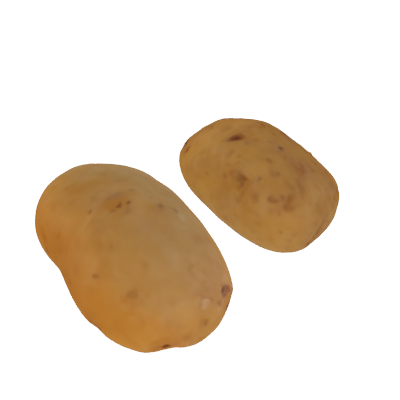} &  
    \includegraphics[width=0.14\linewidth]{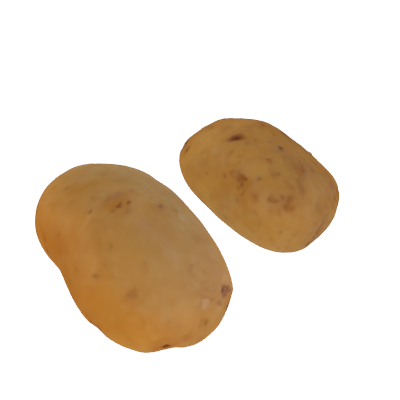}&
    \includegraphics[width=0.27\linewidth]{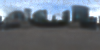} &  
    \includegraphics[width=0.27\linewidth]{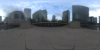} \\
    & \footnotesize PSNR = 31.2 dB &  & \footnotesize RMSE = 0.073 & \\
    \includegraphics[width=0.14\linewidth]{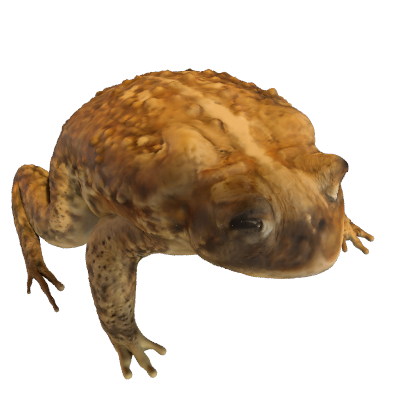} &  
    \includegraphics[width=0.14\linewidth]{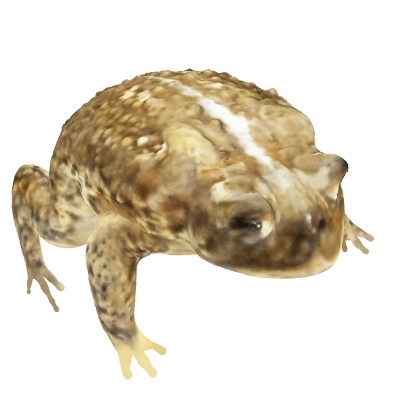} &  
    \includegraphics[width=0.14\linewidth]{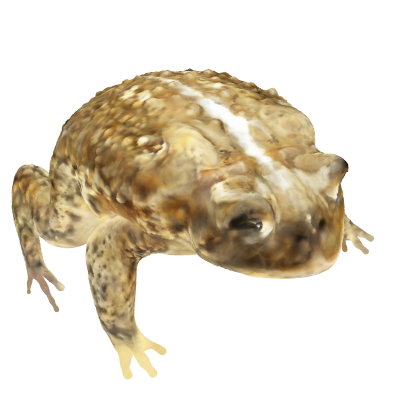} &  
    \includegraphics[width=0.27\linewidth]{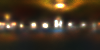} &  
    \includegraphics[width=0.27\linewidth]{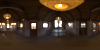} \\
    & \footnotesize PSNR = 26.4 dB & & \footnotesize RMSE = 0.081 & \\
    \includegraphics[width=0.14\linewidth]{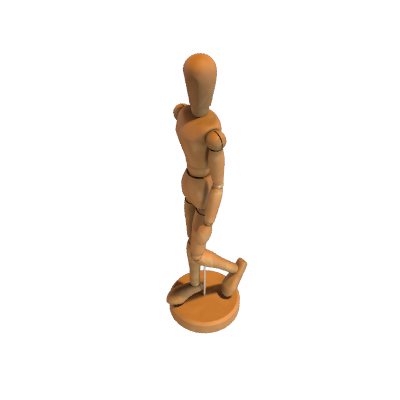} & 
    \includegraphics[width=0.14\linewidth]{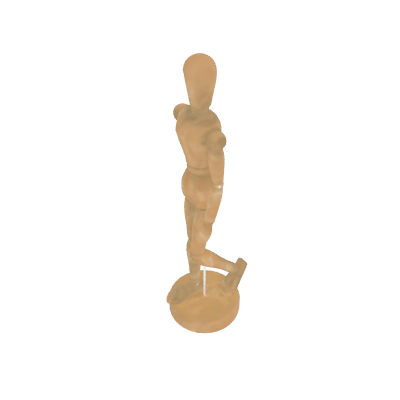} & 
    \includegraphics[width=0.14\linewidth]{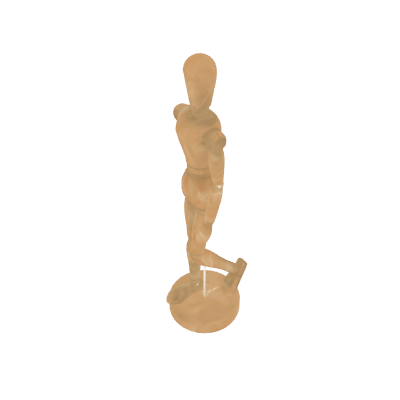} &
    \includegraphics[width=0.27\linewidth]{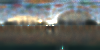} & 
    \includegraphics[width=0.27\linewidth]{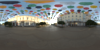} \\ 
    & \footnotesize PSNR = 29.6 dB & & \footnotesize RMSE = 0.093 & \\    
    \scriptsize (a) Sample input image  & \scriptsize (b)  Recovered albedo & \scriptsize (c)  True albedo & \scriptsize (d)  Recovered illumination  & \scriptsize (e)  True illumination
    \end{tabular} 
    \vspace{-0.1in}        
    \caption{The results of our method on three additional diffuse objects. We report the RMSE of each environment map in linear color space but plot the images after tonemapping for better evaluation of the full dynamic range. The albedo PSNR values are reported on object pixels only.}
    \label{fig:moreresults}
\end{figure*}

\subsection{Formulation}

Let $\{I_t\}_{t=1}^{T}$ be a collection of $T$ images of a scene with known camera poses; for every $t$, $I_t$ is an RGB image with spatial size $H\times W$. 
We assume the observed scene ${\cal S}\subset\mathbb{R}^3$ is illuminated by a far-field environment map $L(\lightdir)$ that does not change over time, so the only temporal changes in the incident light field are caused by unobserved light-blockers (called \emph{occluders} hereafter) that move around the scene outside the field of view. We can then write the incident light at any observed position $\bx\in{\cal S}$ as:
\begin{equation}
    L_t(\bx, \lightdir) = L(\lightdir) M_t(\bx, \lightdir) V(\bx, \lightdir)\,,
\end{equation}
where for any $t$, $M_t(\bx, \lightdir)$ is a binary signal with value $0$ for directions blocked by the occluder and $1$ otherwise, and the binary signal $V(\bx, \lightdir)$ models visibility effects that are \emph{internal} to the scene, with value $0$ for directions from $\bx$ that are blocked by other scene elements, and value $1$ otherwise.

Omitting global illumination effects, a pixel $\bu$ corresponding to a surface point $\bx$ with BRDF $f$ viewed from direction $\viewdir$ at time $t$ has color:
\begin{equation} \label{eq:rendering}
    I_t(\bu) = \int_{\mathbb{S}^2} L_t(\bx, \lightdir) f(\bx, \lightdir, \viewdir) (\uvn(\bx)\cdot \lightdir)_+ \, d\lightdir \,,
\end{equation}
where $\uvn(\bx)$ is the surface normal corresponding to $\bx$, and $(\cdot)_+$ clamps negative values to zero.

Given the set of observed images, we would like to recover the environment map $L$, a spatially-varying BRDF $f$ at every point on the surface of the object $\mathcal{S}$, as well as the set of unobserved occluder masks $\{M_t\}_{t=1}^{T}$, one for each observation time $t$. Our goal is therefore to solve the following optimization problem:
\begin{equation} \label{eq:optimization}
    \underset{{\params^{(o)}, \params^{(m)}, \params^{(\ell)}}}{\operatorname{arg\,min}} \sum_{t, \bu} \left\| I_t(\bu) - \mathcal{R}_t\!\left(\bu; \params^{(o)}, \params^{(m)}, \params^{(\ell)} \right) \right\|^2,
\end{equation}
where $\mathcal{R}_t(\bu;\cdot)$ renders the pixel location $\bu$ at time $t$ using the occluder parameters $\params^{(o)}$, material parameters $\params^{(m)}$, and illumination parameters $\params^{(\ell)}$. 

At first glance, it may seem that solving for occluders in addition to materials and illumination should make the resulting inverse problem more difficult. However, as illustrated in Figure~\ref{fig:eigenvalues} and demonstrated by our experimental results, modeling and solving for occluders actually makes the full problem \emph{more} well-conditioned and thus improves recovery of material and illumination.

\section{Method}

We first describe our parameterizations of the unseen occluders, illumination, and materials (Sections~\ref{sec:occluders}--\ref{sec:materials}). Section~\ref{sec:rendering} then relates these parameters to rendered pixel colors, and Section~\ref{sec:optimization} describes how we optimize to solve the inverse problem in Equation~\ref{eq:optimization}. 

We emphasize that our method is designed to exploit external shadows across as many scene-types as possible, and so its only priors are those implicit to our parameterizations. In our experiments, we verify that these relatively weak priors are sufficient for recovering high-fidelity occluder shapes, illumination maps, and material maps.

\begin{figure}
    \centering
    \includegraphics[width=\linewidth]{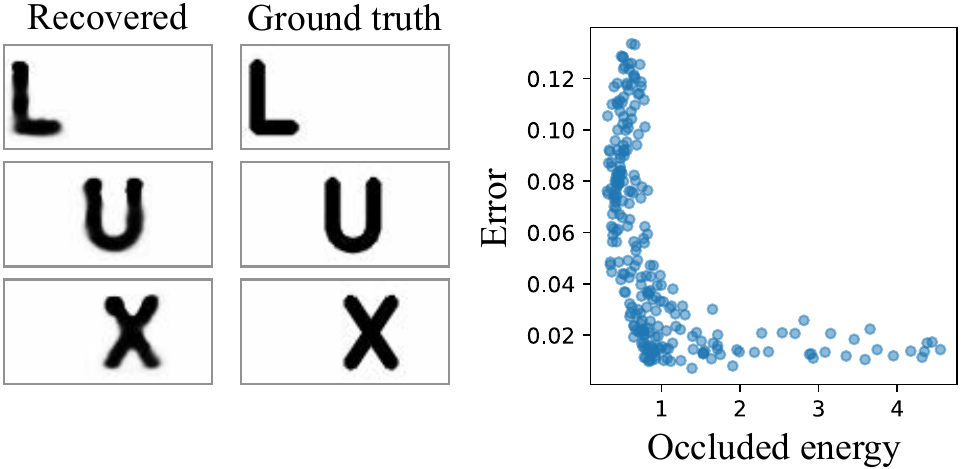}
    \caption{Our occluders extracted from the \emph{potatoes} scene on the top of Figure~\ref{fig:moreresults} resemble the true ones (left). However, as expected, inaccuracies in the recovered masks tend to occur at relatively-dark regions of the environment map: there is a very strong inverse correlation (right) between the mean error in the recovered masks and the amount of illumination energy that they block.}
    \label{fig:iccvmasksenergy}
\end{figure}

\begin{figure*}[t]
    \centering
    \begin{tabular}{@{}c@{\,\,\,}c@{\,\,\,}c@{}}
    \includegraphics[width=0.3\linewidth]{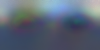} & 
    \includegraphics[width=0.3\linewidth]{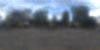} & 
    \includegraphics[width=0.3\linewidth]{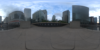} 
    \\
    \footnotesize (a) Recovered illumination (no occluder) & \footnotesize (b) Recovered illumination & \footnotesize (c) True illumination 
    \end{tabular}
    \vspace{-0.12in}
    \caption{In the case of purely Lambertian material (with the \emph{potatoes} geometry), (a) only using self occlusions is insufficient for the recovery of high-frequency content in the illumination. (b) Modeling and estimating occluders recovers significantly higher-quality illumination that closely resembles the ground truth.
    \label{fig:lambert}
    }
\end{figure*}

\subsection{Occluders} \label{sec:occluders}

Define a world coordinate system with origin at the scene's center, and let $\{\radius_t\}_{t=1}^T$ be the radial distances from the origin to each (known) camera's center. Then we model each of the $T$ occluders $M_t$ as an independent binary signal defined on the surface of a finite, radius-$\radius_t$ sphere that is centered at the origin. The value of the binary occluder signal for a shadow ray emitted from surface point $\bx \in \mathcal{S}$ in direction $\lightdir$ at time $t$ can be computed by intersecting the shadow ray with the occluder's spherical shell:
\begin{equation}
    M_t(\bx, \lightdir) = \tilde{M}_t\left(S(\bx, \lightdir, \radius_t)\right)\,
\end{equation}
where $S(\bx, \lightdir, \radius)$ is the intersection of the ray with a sphere of radius $\radius_t$, normalized to have unit length:
\begin{equation}
\resizebox{0.9\linewidth}{!}{$
\displaystyle
S(\bx, \lightdir, \radius) = \frac{\bx + \left(\sqrt{(\bx\cdot\lightdir)^2+\radius^2-\|\bx\|^2} - \bx\cdot\lightdir\right)\lightdir}{\radius}\,.
$}
\end{equation}
Note that there is a single intersection, since we assume that the surface points are always inside the occluder spheres.

Although our formulation models the occluders as binary signals, a binary representation is not well-suited for gradient-based optimization. Instead, we represent the occluders as a continuously-valued function on the sphere in a spherical harmonic basis, mapped using a sigmoid function $\sigma$ to lie in $[0, 1]$:
\begin{equation}
    \tilde{M}_t(\bomega) = \sigma\!\left(\sum_{\ell=0}^{P}\sum_{m=-\ell}^{\ell}a_{t \ell m}Y_\ell^m(\bomega)\right)\,,
\end{equation}
where $\params^{(o)} \eqdef \{a_{t \ell m}\}$ are optimizable coefficients that parameterize the occluder at time $t$, and $P$ is the degree required to span the space of spherical images with the same resolution as our environment illumination (see below)~\cite{driscoll1994computing}. 

Note that for radially symmetric BRDFs, the rendering integral in Equation~\ref{eq:rendering} can be written as a spherical convolution~\cite{driscoll1994computing,ramamoorthi2001spfw}. This makes spherical harmonics a natural choice for representing the occluder signals, since spherical convolutions are diagonal in the basis of spherical harmonics. See our supplement for more details.

\subsection{Environment Illumination}\label{sec:illumination}

We assume that the scene's illumination is distant and can be represented as an environment map. The environment map is parameterized as an image pyramid of size $H'\times W'$ in equirectangular coordinates ($50 \times 100$ in our experiments), with an exponential nonlinearity:
\begin{equation}
    L(\lightdir) = \exp\!\left(\sum_{k=0}^{K-1} a^k L_k(\lightdir)\right)\,,
\end{equation}
where $\params^{(\ell)} \eqdef \{L_k\}_{k=0}^{K-1}$ comprise a coarse $(k=0)$ to fine $(k=K-1)$ representation of the illumination, and we set $a=2$. For each $k$, $L_k(\lightdir)$ is computed by bilinearly interpolating into a grid whose size exponentially increases with $k$. The exponential nonlinearity is helpful for obtaining high dynamic range values in the environment map, as noted, \eg in~\cite{barron2015sirfs}.

Unlike the occluder masks, we find that representing the environment map as a pyramid results in better performance than what is achieved with spherical harmonics. See appendix for additional details.

\subsection{Materials}\label{sec:materials}

We represent the scene's spatially-varying BRDF as the sum of a diffuse and a specular component. We use a Lambertian model for the diffuse term and a GGX microfacet model~\cite{walter2007microfacet} for the specular term. The spatially-varying BRDF parameters are represented by a coordinate-based multi-layer perceptron (MLP) with a positional encoding~\cite{tancik2020fourier} function $\gamma$ and optimizable parameters $\params^{(m)}$:
\begin{equation}
    \brdfparams(\bx) = \text{MLP}\!\left(\gamma(\bx); \params^{(m)}\right)\,.
\end{equation}
Here, $\gamma(\bx)$ is the positionally-encoded position on the object's surface, and $\brdfparams(\bx)\in\R^5$ are BRDF coefficients, consisting of an RGB diffuse albedo, a spatially-varying scalar microfacet roughness, and a spatially-varying scalar specular reflectance at normal incidence (equivalent to a reparameterization of the material's index of refraction). See the supplement for an exact specification of our BRDF model.

\newcommand{\frogtop}{0.9in}
\begin{figure}[t!]
    \centering
    \begin{tabular}{@{}c@{\,\,\,}c@{\,\,\,}c@{\,\,\,}c@{}}
    \includegraphics[trim=0in 0.3in 0in \frogtop, clip, width=0.325\linewidth]{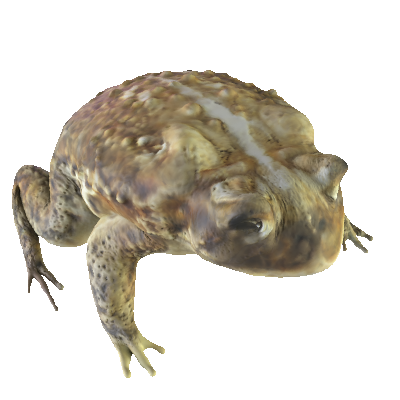} & 
    \includegraphics[trim=0in 0.3in 0in \frogtop, clip, width=0.325\linewidth]{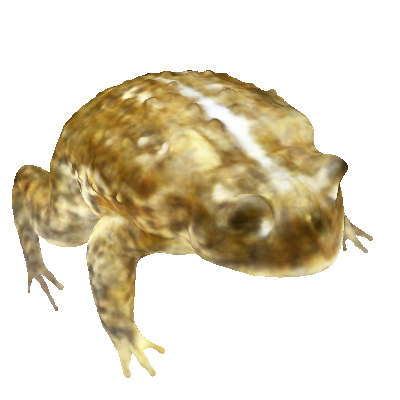} &
    \includegraphics[trim=0in 0.3in 0in \frogtop, clip, width=0.325\linewidth]{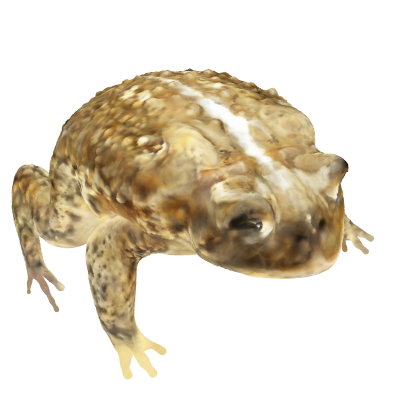} \\[-0.75ex]
    \footnotesize (a) Sample input image  & \footnotesize (b) Recovered albedo &  \footnotesize (c) True albedo \\[1ex]
    \end{tabular}
    \begin{tabular}{@{}c@{\,\,\,}c@{\,\,\,}c@{}}
    \includegraphics[width=0.495\linewidth]{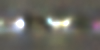} &
    \includegraphics[width=0.495\linewidth]{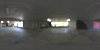} \\[-0.75ex]
    \footnotesize (d) Recovered illumination &
    \footnotesize (e) True illumination 
    \end{tabular} 
    \vspace{-0.1in}
    \caption{We are able to recover illumination even when geometry is unknown by first optimizing a volumetric representation of geometry using a NeRF-based method. Despite this inaccurate proxy geometry, our method still recovers plausible (albeit blurry) illumination (d) and albedo (b). 
    \label{fig:ngpresults}
    }
\end{figure}

\subsection{Rendering}\label{sec:rendering}

We render the occluders, materials, and illumination, \ie the rendering operator $\mathcal{R}_t$ from Problem~\ref{eq:optimization}, by approximating the integral in Equation~\ref{eq:rendering} using Monte Carlo techniques. We use a standard multiple importance sampler~\cite{veach1995optimally} based on the illumination and material model. We model shadows cast by the occluders and self-occlusions by the object itself, but for efficiency and due to our focus on diffuse objects, we neglect lower-order effects and global illumination. See the supplemental material for a full description of our rendering engine.

\begin{figure*}[t!]
    \centering
    \begin{tabular}{@{}c@{\,\,\,}c@{\,\,\,}c@{\,\,\,}c@{}}
    \includegraphics[trim=0in 0.3in 0in \frogtop, clip, width=0.245\linewidth]{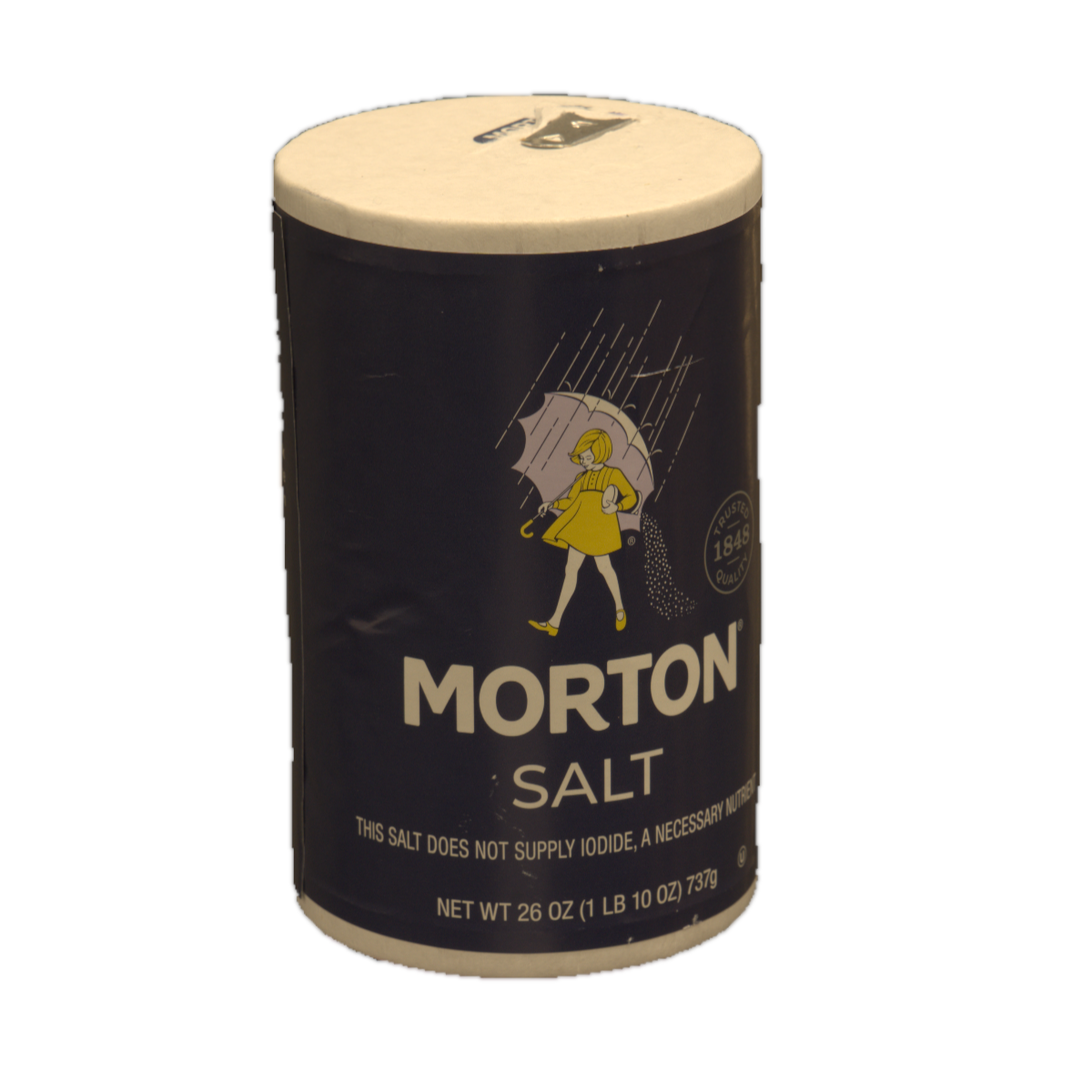} & 
    \includegraphics[trim=0in 0.3in 0in \frogtop, clip, width=0.245\linewidth]{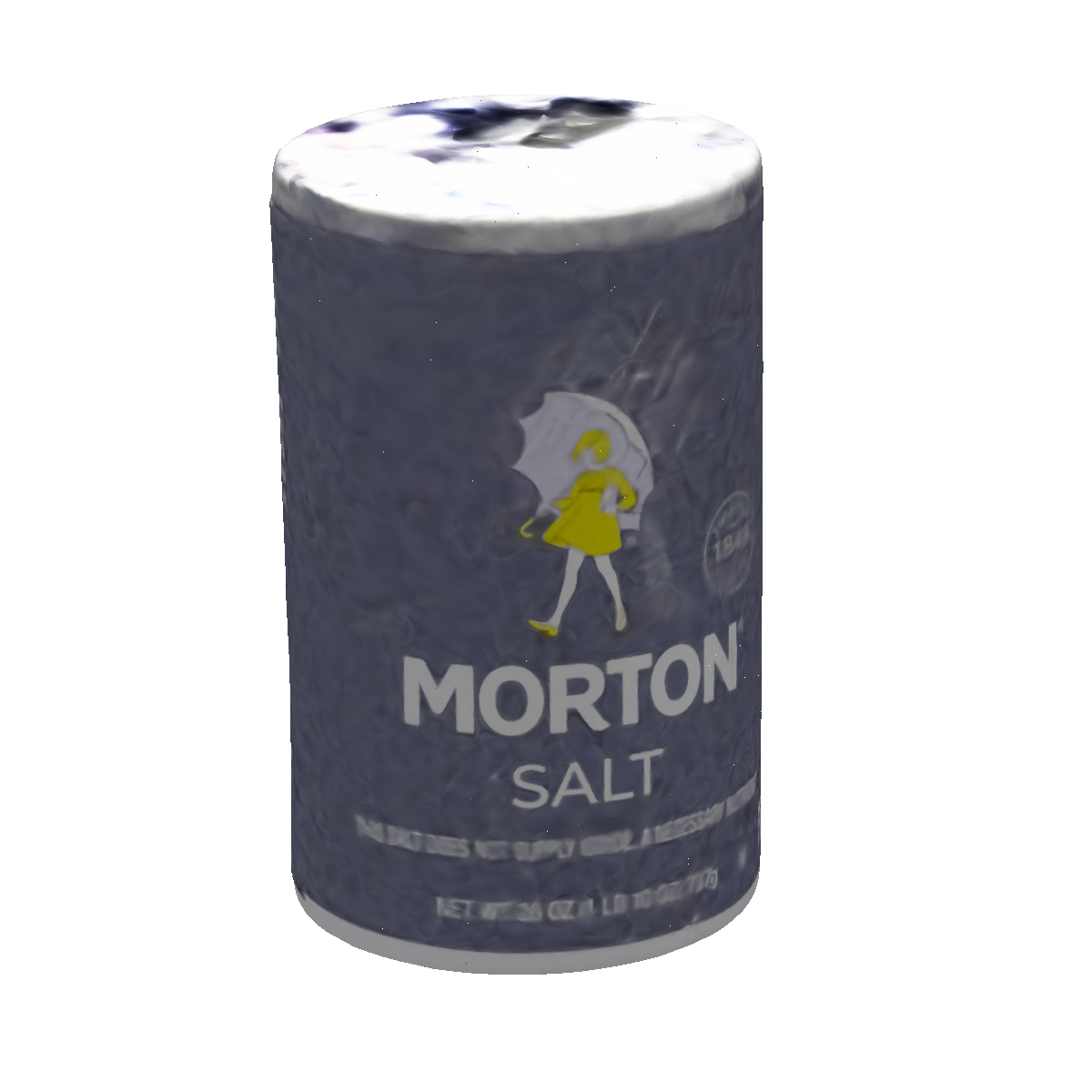} &
    \includegraphics[trim=0in 0.3in 0in \frogtop, clip, width=0.245\linewidth]{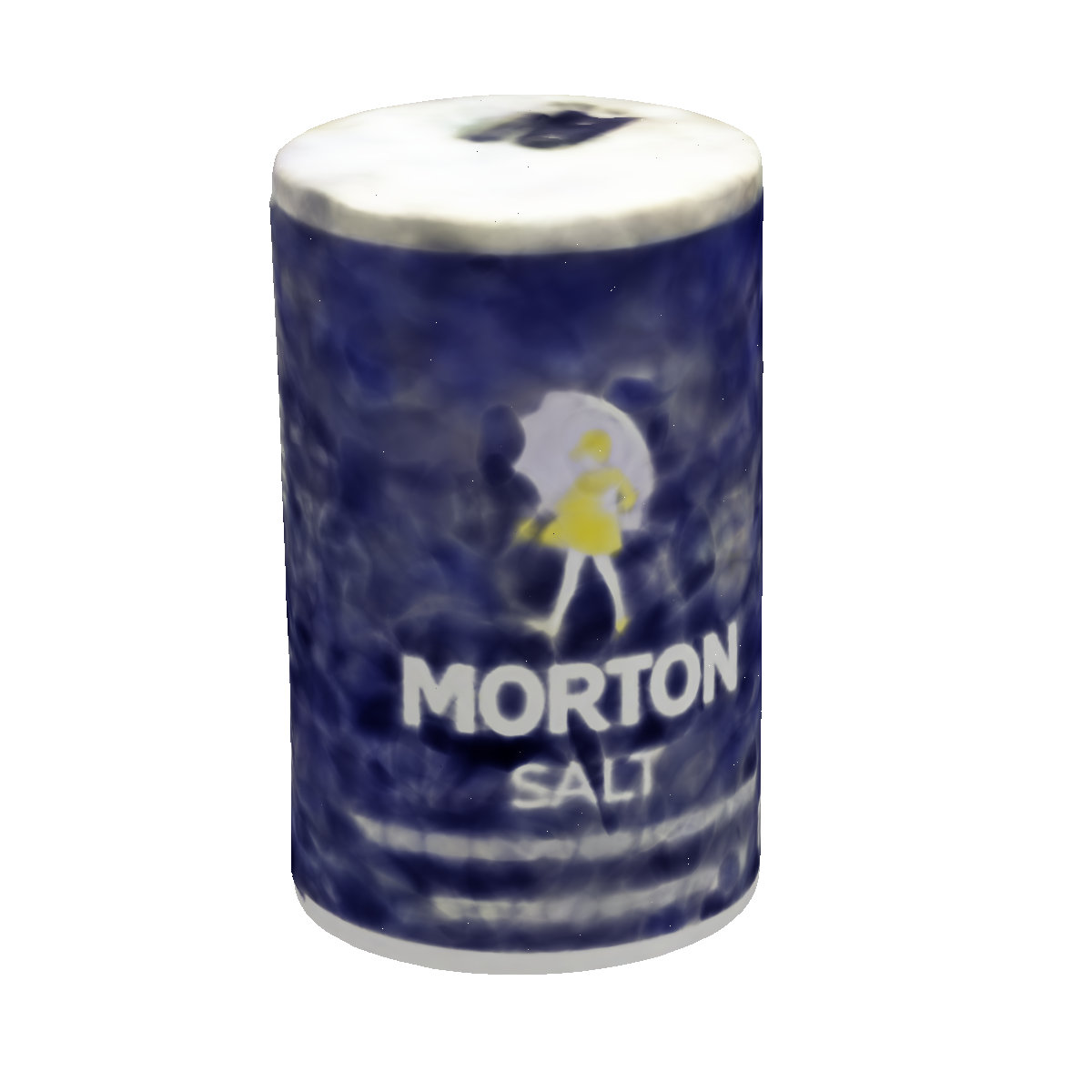} &
    \includegraphics[trim=0in 0.3in 0in \frogtop, clip, width=0.245\linewidth]{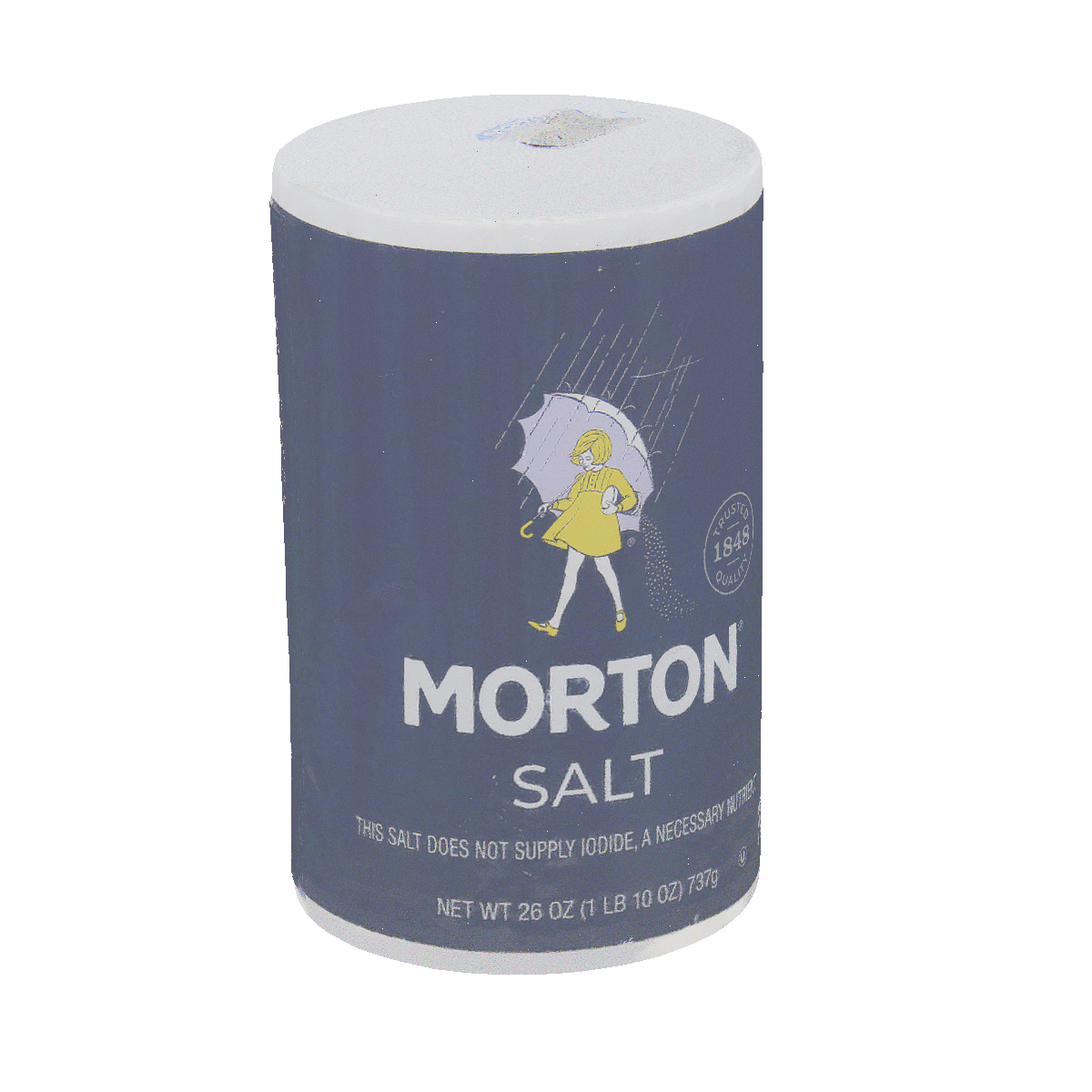} \\[-0.75ex]
    & \footnotesize PSNR = 19.3 dB &  \footnotesize PSNR = 14.4 dB & \\    
    \footnotesize (a) Sample input image  & \footnotesize (b) Recovered albedo & \footnotesize (c) Recovered albedo (no occluder) & \footnotesize (d) True albedo \\[1ex]
    \end{tabular}
    \begin{tabular}{@{}c@{\,\,\,}c@{\,\,\,}c@{}}
    \includegraphics[width=0.325\linewidth]{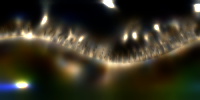} &
    \includegraphics[width=0.325\linewidth]{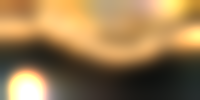} &
    \includegraphics[width=0.325\linewidth]{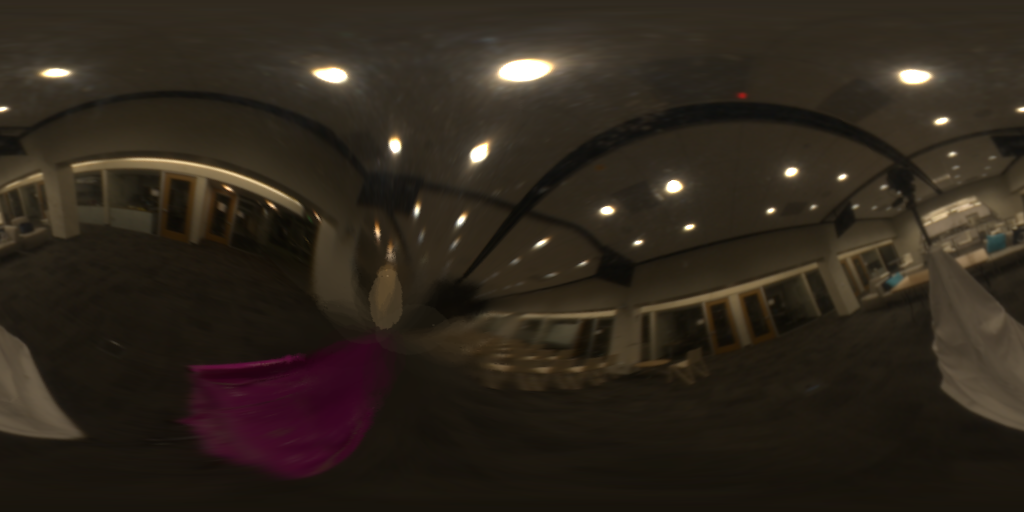} \\[-0.75ex]
    \footnotesize RMSE = 0.035 &  \footnotesize RMSE = 0.087 & \\        
    \footnotesize (e) Recovered illumination & \footnotesize (f) Recovered illumination (no occluder) & \footnotesize (g) \textbf{Warped} true illumination 
    \end{tabular} 
    \vspace{-0.1in}
    \caption{Results on the captured \texttt{salt004} scene from Stanford-ORB~\cite{kuang2023stanford}. Note that the ``warped true illumination'' environment map (g) was captured by a light probe that was not co-located with the object, any may therefore be significantly warped. The recovered RMSE reported is computed with respect to the average of all environment maps provided with the dataset rotated the same coordinate frame.
    \label{fig:orb}
    }
\end{figure*}

\subsection{Optimization}\label{sec:optimization}

We optimize the objective in Equation~\ref{eq:optimization} using the $L_2$ error between rendered values and ground truth ones. We use Adam~\cite{kingma2015adam} to optimize all three components of our model, with a learning rate of $3\cdot 10^{-3}$ for the environment map and materials, and a learning rate of $1$ for the occluders.

In each iteration, we compute the $L_2$ loss using a batch size of $2^{16}$ pixels. In order to avoid bias in the gradient updates to our model's parameters, we use the same approach as~\cite{deng2022reconstructing}: we render every pixel value twice using independent samples and multiply the per-channel deviations of both from the ground truth to get the expected loss value:
\begin{equation}
    \mathcal{L} = \sum_{\bu}\left(\tilde{I}_t^{(1)}(\bu) - I_t(\bu)\right) \cdot \left(\tilde{I}_t^{(2)}(\bu) - I_t(\bu)\right)\,,
\end{equation}
where $\tilde{I}_t^{(1)}(\bu)$ and $\tilde{I}_t^{(2)}(\bu)$ are the two independently-rendered pixel values.

\begin{figure*}[t!]
    \centering
    \begin{tabular}{@{}ccc@{}}
    \begin{overpic}[width=0.315\linewidth]{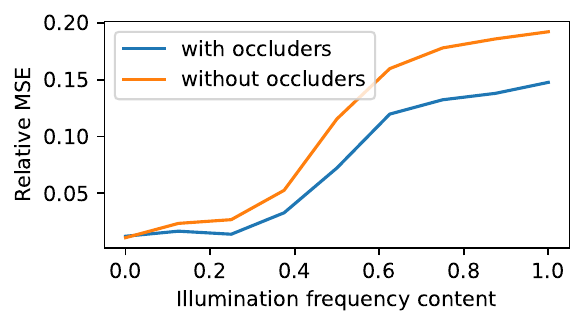}
    \put(2,3){\scriptsize (a)}
    \end{overpic}
    &
    \begin{overpic}[width=0.315\linewidth]{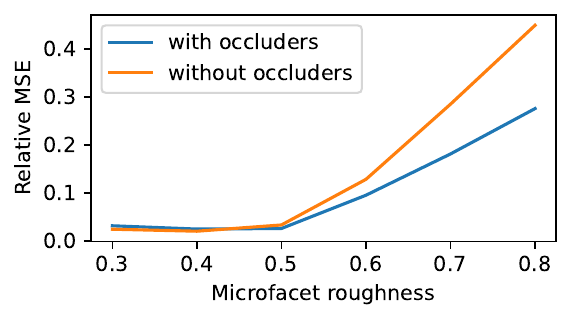}
    \put(2,3){\scriptsize (b)}
    \end{overpic}
    &
    \begin{overpic}[width=0.315\linewidth]{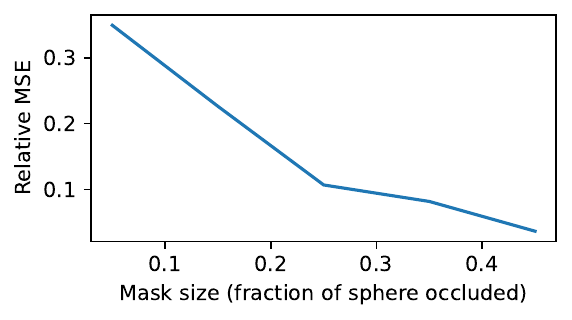}
    \put(2,3){\scriptsize (c)}
    \end{overpic}     
    \vspace{-0.1in}
    \end{tabular}
    \caption{The signal provided by unintended shadows cast by unseen occluders improves the quality of recovered environment illumination. Here, we plot the relative MSE for the recovered environment maps under two scenarios: (blue) using images rendered with unobserved occluders and jointly estimating materials, illumination and occluder shape and (orange) using images rendered without occluders and only estimating materials and illumination (this is the problem setting considered by Swedish \etal~\cite{swedish2021objects}). The cue of unintended shadows consistently improves the quality of estimated illumination across varying (a) illumination frequency content, (b) object material roughness, and (c) mask sizes. Additionally, we show that larger occluders that block more light improve the reconstructed illumination quality.}
    \label{fig:ablations}
\end{figure*}

\begin{figure}
    \centering
    \begin{tabular}{@{}c@{\,\,}c@{}}
    \includegraphics[width=0.48\linewidth]{figures/results/toad/toad_envmap.png} & \includegraphics[width=0.48\linewidth]{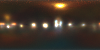}
    \\
    \includegraphics[width=0.48\linewidth]{figures/results/potato/potato_envmap.png} & \includegraphics[width=0.48\linewidth]{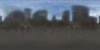} \\[-0.75ex]
    \scriptsize (a) Illumination (pyramid) & \scriptsize (b) Illumination (direct)

    \end{tabular}
    \vspace{-0.1in}
    \caption{Results from the experiment in Figure~\ref{fig:moreresults}, where the pyramid environment map has been replaced with a direct optimization of its (pre-exponentiated) values. Directly optimizing the mask values instead of using spherical harmonics also results in worse reconstructions (see Table~\ref{tab:ablation}).}
    \label{fig:representationablation}    
\end{figure}

\section{Experiments}

We implement our entire rendering and optimization pipeline in JAX~\cite{jax2018github} and run each of our experiments on 8 NVIDIA V100 GPUs. We evaluate our method on two sources of data: synthetic objects, and captured data from the Stanford-ORB dataset~\cite{kuang2023stanford}.
Our synthetic results were generated using $256$ input images of size $400\times 400$, although our method works similarly well with as few as 32 captures (see supplement for the effect of the number of images).

\boldstart{Estimating illumination, materials, and occluders.} Our method can effectively recover environment illumination, spatially-varying material parameters, and the shapes of unseen lighting occluders. We first investigate our algorithm's performance using rendered images of a variety of diffuse objects (roughness $0.6$, see BRDF model in appendix) with known geometry. The illumination and albedos recovered by our model are visualized with their true values in Figures~\ref{fig:teaser} and~\ref{fig:moreresults}. Figure~\ref{fig:lambert} shows our reconstruction for purely Lambertian materials, compared with the reconstruction from the same number of images \emph{without} unobserved occluders (but with self-occlusions), similar to the scenario investigated by Swedish~\etal~\cite{swedish2021objects}. For Lambertian objects, reconstruction is severely ill-conditioned even when self-occlusions are present, which results in significantly blurrier reconstructions. Figure~\ref{fig:iccvmasksenergy} validates that our recovered occluder shapes are accurate, especially for occluders that block more of the illumination energy. 

\boldstart{Captured data.} We apply our method to scenes from the Stanford-ORB dataset~\cite{kuang2023stanford}. Each object in the dataset is placed on a platform and captured from multiple directions. As the camera rig and photographer move around the scene, the photographer is hidden under white cloth below the camera. The images are provided along with a scanned mesh which we use for geometry.

Despite not being designed for our task, Figure~\ref{fig:orb} shows our recovered environment map obtained using the $66$ training views in the \texttt{salt004} scene, which features a rough cylindrical object. Note that the dataset only provides an estimated illumination obtained by a light probe placed at a large distance from the object, meaning there is a significant unknown non-rigid warp between the true incident light and the image in Figure~\ref{fig:orb}g labeled ``warped true illumination''. See supplement for additional results.

\boldstart{Without known geometry.} Figure~\ref{fig:ngpresults} presents preliminary evidence that our method can be used to recover lighting and materials even when the object geometry is not known. In this experiment, we first recover an estimate of object geometry using a Neural Radiance Field (NeRF)~\cite{mildenhall2020nerf}-based model augmented using the orientation loss from Verbin~\etal~\cite{verbin2022refnerf}, and then use this (potentially-imprecise) proxy geometry in our inverse rendering optimization pipeline. 
Please refer to the supplemental materials for a detailed description of our NeRF-based model.

\boldstart{Object self-occlusion does not contain enough signal.} Figure~\ref{fig:ablations} quantitatively demonstrates that only using the signal provided by object self-occlusions (``without occluders'' plot in orange) cannot recover illumination as accurately as our method, which leverages the cue of unintended shadows (``with occluders'' plot in blue). In particular, exploiting unintended shadows is increasingly important as the illumination contains higher frequencies (Figure~\ref{fig:ablations}a) and as the object material becomes increasingly diffuse (Figure~\ref{fig:ablations}b).

\begin{table}
    \centering
    \resizebox{\linewidth}{!}{
    \begin{tabular}{@{}l|c@{\,\,\,}c@{\,\,\,}c@{\,\,\,}c@{}}
    & \textit{chair} & \textit{mannequin} & \textit{potatoes} & \textit{toad} \\ \hline
    Direct spherical harmonics     & 0.831  &  0.654     & 0.997    & 0.943 \\
    Direct environment map         & 0.042 &  0.071     & 0.015     & 0.073 \\ \hline
    Ours                           & \textbf{0.038} &  \textbf{0.035}     & \textbf{0.011}     & \textbf{0.046}
    \end{tabular}
    }
    \vspace{-0.1in}
    \caption{Errors (RMSE) on all four of our scenes for our full model compared with our model without the spherical harmonic occluder parameterization, and without the pyramid environment parameterization. While the pyramid representation of the environment map promotes smoothness (see Figure~\ref{fig:representationablation}), we find that representing the masks using spherical harmonics is critical for our method's success.
    }
    \label{tab:ablation}
\end{table}

\boldstart{Larger occluders improve illumination recovery.} Figure~\ref{fig:ablations}c shows that increasing the size of the unseen occluders improves the accuracy of the recovered environment maps. Increasing the occluder size effectively improves the problem's conditioning by creating a larger variation in the incident lighting across points on the object.

\paragraph{Parameterization ablation study.}

Table~\ref{tab:ablation} and Figure~\ref{fig:representationablation} show that representing the environment map as an image pyramid and the image masks with spherical harmonics, both perform better than simply optimizing their respective element values directly. See supplement for more details.

\section{Discussion}

Even though unintended shadows are common in most real-world capture scenarios, they are often treated as outlier data. In this paper, we showed how to explicitly leverage these shadows as a signal to improve the quality of recovered lighting and materials in particularly challenging scenarios, such as objects made of diffuse materials. Furthermore, we have shown that our algorithm can be used with approximate geometry reconstructed by view synthesis techniques, and that it can produce promising results on real captures, despite not being designed for this type of data which does not satisfy many of our model's assumptions. We believe that this work makes important first steps towards a general inverse rendering algorithm that can recover geometry, materials, and illuminations from images captured under realistic conditions.

\boldstart{Acknowledgements.} Our sincere thanks to Delio Vicini, Rick Szeliski, Aleksander Hołyński, and Guy Satat for helpful discussions and suggestions, and to David Salesin, Janne Kontkanen, and Lior Yariv for their help in improving the manuscript before submission.

{\small
\bibliographystyle{ieeenat_fullname}
\bibliography{main}
}

\clearpage
\setcounter{page}{1}
\setcounter{section}{0}
\setcounter{equation}{0}
\setcounter{figure}{0}
\setcounter{table}{0}
\maketitlesupplementary

\renewcommand{\theequation}{S\arabic{equation}}
\renewcommand{\thefigure}{S\arabic{figure}}
\renewcommand{\thetable}{S\arabic{table}}
\renewcommand{\thesection}{S\arabic{section}}


\section{Additional results}

\begin{figure*}[t]
    \centering
    \begin{tabular}{@{}c@{\,\,}c@{\,\,}c@{\,\,}@{\,}c@{\,\,}c@{}}
    \includegraphics[width=0.14\linewidth]{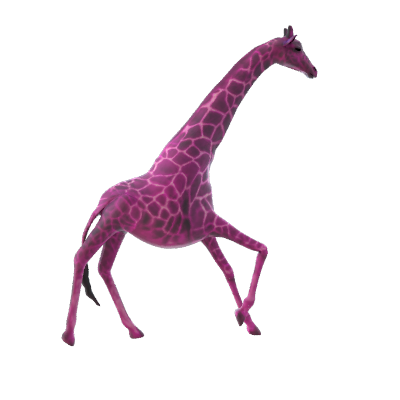} & 
    \includegraphics[width=0.14\linewidth]{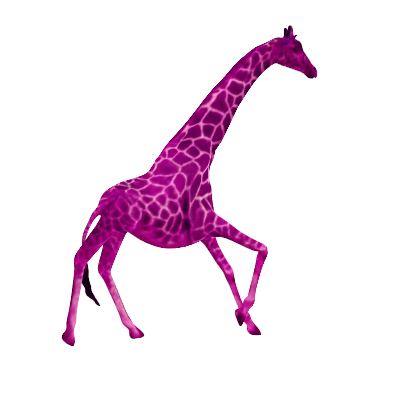} &  
    \includegraphics[width=0.14\linewidth]{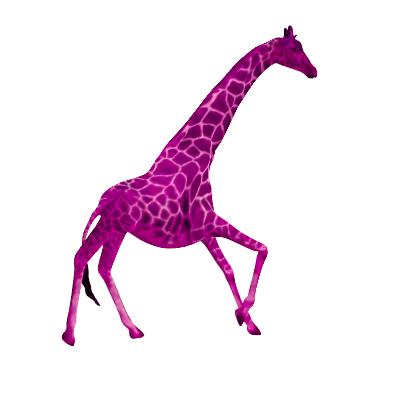}&
    \includegraphics[width=0.27\linewidth]{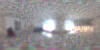} &  
    \includegraphics[width=0.27\linewidth]{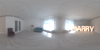} \\
    & \footnotesize PSNR = 32.3 dB &  & \footnotesize RMSE = 0.051 & \\
    \includegraphics[width=0.14\linewidth]{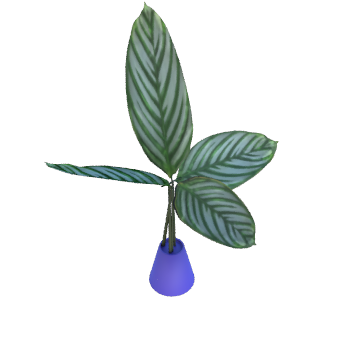} & 
    \includegraphics[width=0.14\linewidth]{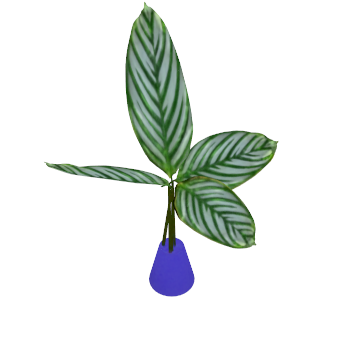} &  
    \includegraphics[width=0.14\linewidth]{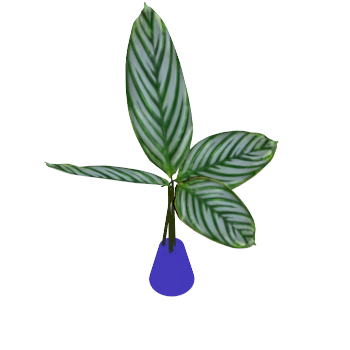}&
    \includegraphics[width=0.27\linewidth]{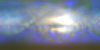} &  
    \includegraphics[width=0.27\linewidth]{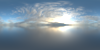} \\
    & \footnotesize PSNR = 32.2 dB &  & \footnotesize RMSE = 0.118 & \\
    \scriptsize (a) Sample input image  & \scriptsize (b)  Recovered albedo & \scriptsize (c)  True albedo & \scriptsize (d)  Recovered illumination  & \scriptsize (e)  True illumination
    \end{tabular} 
    \vspace{-0.1in}
    \caption{Additional results on diffuse objects, similar to Figures~1 and~4 of the main paper. We report the RMSE of each environment map in linear color space but plot the images after tonemapping for better evaluation of the full dynamic range. The albedo PSNR values are reported on object pixels only.}
    \label{fig:suppsynthetic}
\end{figure*}

\begin{figure*}[t!]
    \centering
    \begin{tabular}{@{}c@{\,\,\,}c@{\,\,\,}c@{\,\,\,}c@{}}
    \includegraphics[trim=0in 0.3in 0in 0in, clip, width=0.245\linewidth]{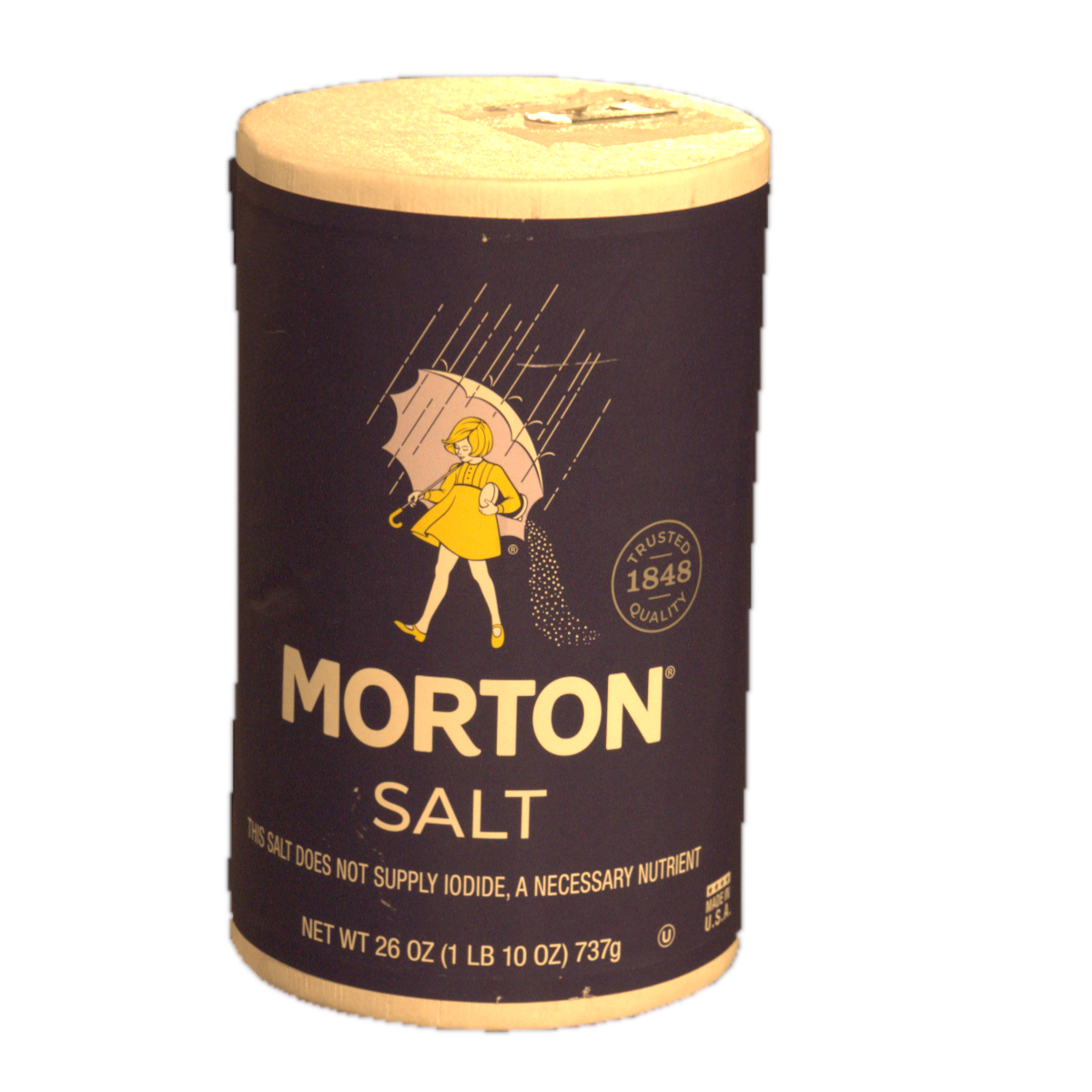} & 
    \includegraphics[trim=0in 0.3in 0in 0in, clip, width=0.245\linewidth]{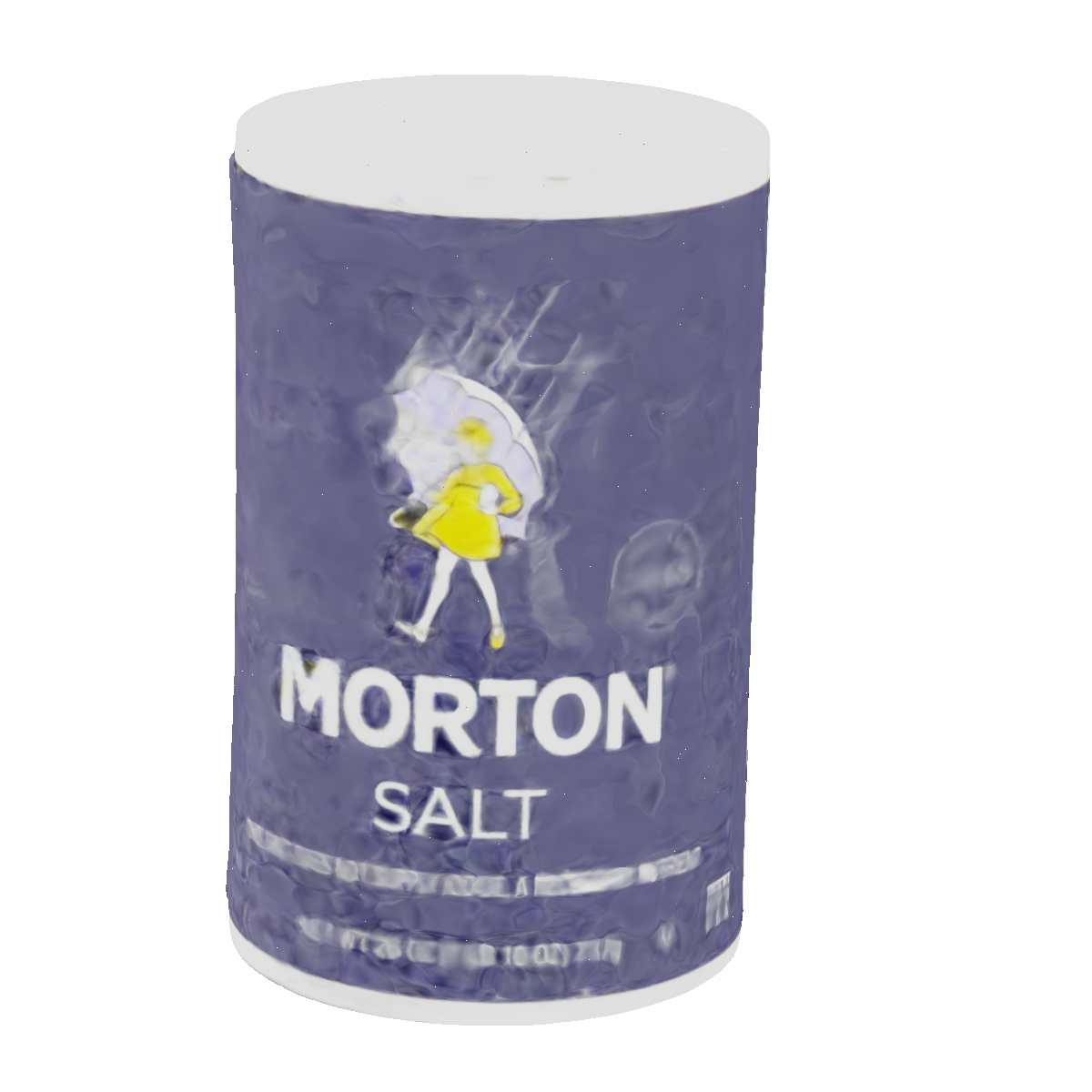} &
    \includegraphics[trim=0in 0.3in 0in 0in, clip, width=0.245\linewidth]{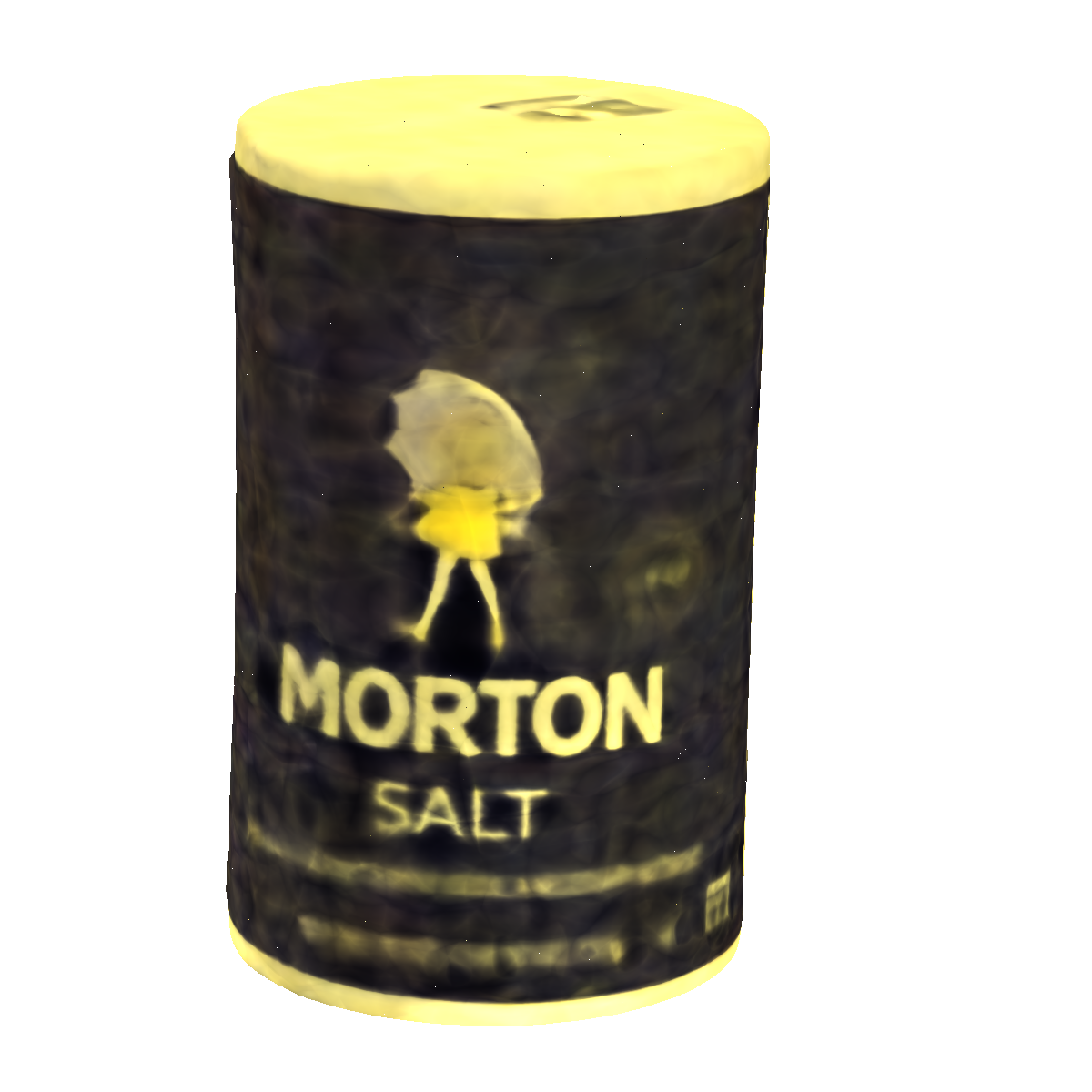} &
    \includegraphics[trim=0in 0.3in 0in 0in, clip, width=0.245\linewidth]{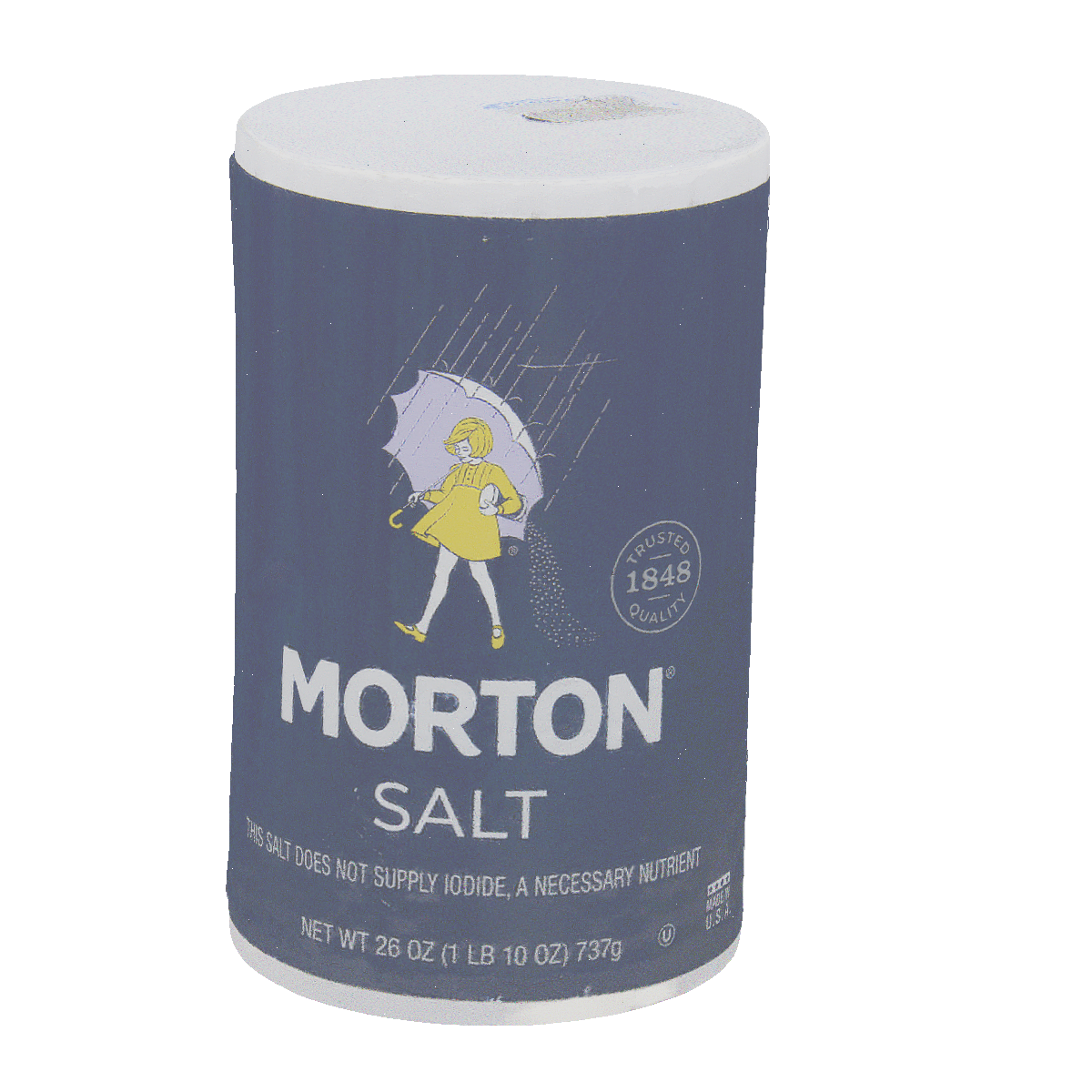} \\[-0.75ex]
    & \footnotesize PSNR = 18.9 dB &  \footnotesize PSNR = 13.5 dB & \\    
    \footnotesize (a) Sample input image  & \footnotesize (b) Recovered albedo & \footnotesize (c) Recovered albedo (no occluder) & \footnotesize (d) True albedo \\[1ex]
    \end{tabular}
    \begin{tabular}{@{}c@{\,\,\,}c@{\,\,\,}c@{}}
    \includegraphics[width=0.325\linewidth]{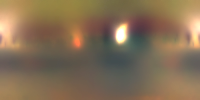} &
    \includegraphics[width=0.325\linewidth]{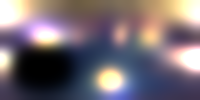} &
    \includegraphics[width=0.325\linewidth]{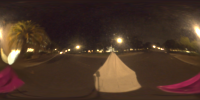} \\[-0.75ex]
    \footnotesize RMSE = 0.039 &  \footnotesize RMSE = 0.112 & \\        
    \footnotesize (e) Recovered illumination & \footnotesize (f) Recovered illumination (no occluder) & \footnotesize (g) \textbf{Warped} true illumination 
    \end{tabular} 
    \vspace{-0.1in}
    \caption{Results on the captured \texttt{salt007} scene from Stanford-ORB~\cite{kuang2023stanford}, extending Figure~8 of the main paper. Note that the ``warped true illumination'' environment map (g) was captured by a light probe that was not co-located with the object, any may therefore be significantly warped. The recovered RMSE reported is computed with respect to the average of all environment maps provided with the dataset rotated the same coordinate frame, see text and Figure~\ref{fig:envmapblur}.
    \label{fig:supporb}
    }
\end{figure*}

Figure~\ref{fig:suppsynthetic} contains additional synthetic results, similar to Figures~1 and~4 of the main paper, and Figure~\ref{fig:supporb} contains additional real results from the Stanford-ORB dataset~\cite{kuang2023stanford}, similar to Figure~8 from the main paper. As explained in the paper, the recovered illumination estimated by our model corresponds to the incident light at the object, which is typically extremely warped relative to the collection of environment maps provided by the dataset as ground truth, since those are obtained using a light probe placed at different points in the scene. We therefore compute our quality metrics relative to the mean environment map obtained from these differently-placed probes, blurred using a (normalized) spherical Gaussian kernel $p(\theta, \phi) \propto \exp(\kappa \cos\theta)$ (we set $\kappa = 100$) to fuse the misaligned light sources. See Figure~\ref{fig:envmapblur} for an example of this process.

\begin{figure*}
\centering
\includegraphics[width=\linewidth]{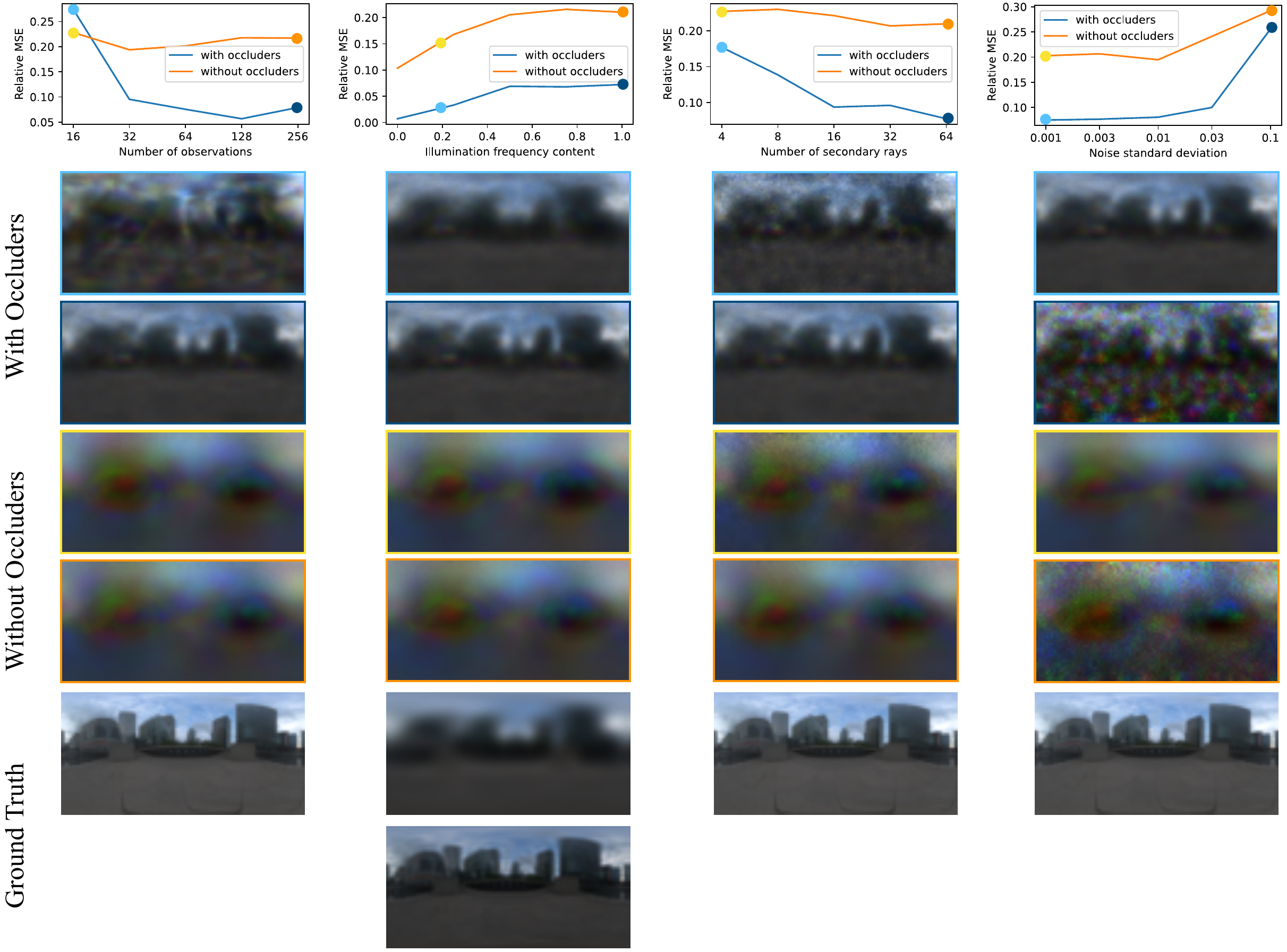}
\caption{The signal provided by unintended shadows cast by unseen occluders improves the quality of recovered environment illumination. This is an expansion of Figure~8 in the main paper, but for an object with purely Lambertian material (we use the \emph{potatoes} scene for these experiments). Here, we plot the relative MSE for the recovered environment maps under two scenarios: (blue) using images rendered with unobserved occluders and jointly estimating materials, illumination and occluder shape; and (orange) using images rendered without occluders and only estimating materials and illumination (this is similar to the problem setting considered by Swedish \etal~\cite{swedish2021objects}, but optimized using our method). The cue of unintended shadows consistently improves the quality of estimated illumination across varying (first column) number of observed images, (second column) illumination frequency content, (third column) number of secondary rays traced, (fewer secondary rays causes increased Monte Carlo noise), and (fourth column) additive Gaussian noise. We display recovered environment maps corresponding to two points on each plot.}
\label{fig:suppplots}
\end{figure*}

Figure~\ref{fig:suppplots} provides additional evidence that the signal provided by unintended shadows improves the recovery of environment illumination, expanding on~Figure 9 in the main paper.

\begin{figure*}
    \centering
    \begin{tabular}{@{}c@{\,}c@{\,}c@{}}
    \includegraphics[width=0.3\linewidth]{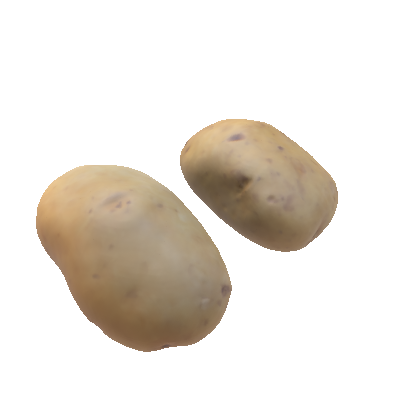} & 
    \includegraphics[width=0.3\linewidth]{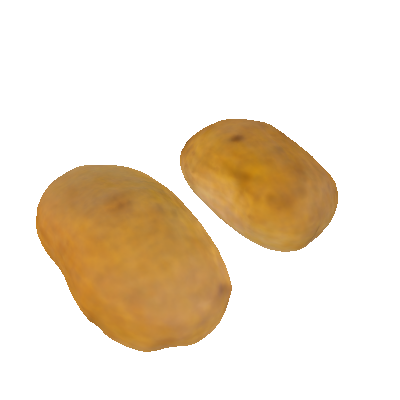} &
    \includegraphics[width=0.3\linewidth]{figures/results/potato/potato_albedo_gt.png} \\
    \scriptsize (a) Sample input image  & \scriptsize (b) Recovered albedo & \scriptsize (c) True albedo \\
    \end{tabular}
    \begin{tabular}{@{}c@{\,}c@{}}
    \includegraphics[width=0.4\linewidth]{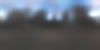} &
    \includegraphics[width=0.4\linewidth]{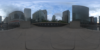} \\
    \scriptsize (e) Recovered illumination & \scriptsize (g) True illumination 
    \end{tabular} 
    \vspace{-0.1in}
    \caption{Expansion of Figure~7 in the main paper. We are able to recover illumination even when geometry is unknown by first optimizing a volumetric representation of geometry using a NeRF-based method. Despite this inaccurate proxy geometry, our method still recovers plausible (albeit blurry) illumination (e) and albedo (b).
    \label{fig:suppngpresults}
    }
\end{figure*}

Figure~\ref{fig:suppngpresults} contains an additional result for recovered illumination and material properties when the object geometry is not given, as in Figure~7 of our main paper.

\paragraph{Relighting.}

Figure~\ref{fig:relitpotato} shows that our recovered albedo enables convincing relighting results. See our supplemental video for rendered video results.

\paragraph{Non-diffuse results.}

Figure~\ref{fig:shinypotato} shows the results obtained by our method when applied to the same microfacet BRDF model from our main paper, but with a roughness value of $0.2$ (see Section~\ref{sec:brdf}).

\section{BRDF model} \label{sec:brdf}

The experiments in the paper and supplement were performed using either a standard Lambertian BRDF (where specified) or a BRDF based on Unreal Engine's version of GGX~\cite{karis2013real,walter2007microfacet}:
\begin{align}
    &\!\!\!\!\!\!\! f(\bx, \lightdir, \viewdir) = \nonumber \\
    & \frac{1}{\pi}\rho(\bx) (1-F(\uvn\cdot\lightdir; \fzero))(1-F(\halfvec\cdot\lightdir, \fzero))(\uvn\cdot\lightdir)_+ \nonumber \\
    +& \frac{D(\uvn\cdot\halfvec; \alpha)F(\halfvec\cdot\lightdir; \fzero)G(\uvn\cdot\viewdir, \uvn\cdot\lightdir; \alpha)}{4(\uvn\cdot\lightdir)_+(\uvn\cdot\viewdir)_+}\,,
\end{align}
where:
\begin{align}
    F(\cos\theta; \fzero) &= \fzero + (1 - \fzero) (1-\cos\theta)^5\,, \\
    D(\cos\theta; \alpha) &= \frac{\alpha^2}{\pi(1 + (\alpha^2-1)\cos^2\theta)^2}\,, \label{eq:trowbridge} \\
    G(\cos\theta_i, \cos\theta_o; \alpha) &= g(\cos\theta_i; \alpha)g(\cos\theta_o; \alpha)\,, \\
    g(\cos\theta; \alpha) &= \frac{\cos\theta}{k(\alpha)+(1-k(\alpha))\cos\theta}\,, \\
    k(\alpha) &= \frac{(\alpha+1)^2}{8}\,, \\
    \halfvec &= \frac{\lightdir + \viewdir}{\|\lightdir + \viewdir\|}\, \label{eq:halfvec}.
\end{align}
We omit the positional dependence of the surface normal vector $\uvn$ in the point $\bx$ on the surface.

The $5$ parameters describing the BRDF at a given location $\bx$ are therefore the RGB albedo $\brho$, the microfacet roughness $\alpha$, and the specular reflectance at normal incidence $\fzero$ which determines the strength of the Fresnel factor $F$.

\section{Full Description of Renderer}

We describe our renderer's approximation of the integral in Equation~2 of the main paper. Our renderer follows well-established practices that are readily documented in public textbooks such as PBRT~\cite{PBRT3e}.

In order to render a specific pixel, we begin by choosing a random ray from the camera center through the square footprint of the pixel. The intersection point of the ray with the object's mesh is then computed, as well as the normal vector at that point (interpolated from per-vertex normals). Then, we use multiply importance sampling, and randomly sample $s_\ell + s_m = 1024$ rays from this intersection point, with $s_\ell = 512$ sampled from the lighting distribution, and $s_m = 512$ sampled from the material distribution. The lighting distribution is simply a piecewise-constant distribution proportional to the average environment map values taken across the three channels, weighted by the Jacobian of the parameterization, $\sin\theta$:
\begin{equation} \label{eq:pl}
    p^{(\ell)}(\bomega) \propto \sum_{c\in\{R, G, B\}}\tilde{L}_{c}(\bomega)\sin\theta,
\end{equation}
where $\tilde{L}_c$ is the $c$-th channel of the environment map obtained using nearest-neighbor interpolation, and $\theta$ is the elevation angle corresponding to $\bomega$, \ie $\sin\theta = \sqrt{1-(\bomega\cdot\hat{\mathbf{z}})^2}$, with $\hat{\mathbf{z}}$ denoting the unit vector in the $z$ direction.

The material distribution is simply the Trowbridge-Reitz normal distribution function reparameterized for sampling incoming light directions:
\begin{equation} \label{eq:pm}
    p^{(m)}(\bomega) = D(\uvn\cdot\halfvec; \alpha)\frac{|\halfvec\cdot\hat{\mathbf{z}}|}{4(\halfvec\cdot\viewdir)_+}\,,
\end{equation}
where $\halfvec$ is the half-vector defined in Equation~\ref{eq:halfvec}, and $D$ is the Trowbridge-Reitz distribution with roughness parameter $\alpha$, as defined in Equation~\ref{eq:trowbridge}.

The two sets of samples are combined using the power heuristic~\cite{veach1995optimally}:
\begin{align} \label{eq:importancesampling}
    C(\bx, \bomega^{(j)}, \viewdir) =  & L_t\!\left(\bomega^{(j)}\right)f_{\bx}\!\left(\bomega^{(j)}, \viewdir\right)\!\left(\uvn(\bx)\cdot\bomega^{(j)}\right)_+ \nonumber \\
    \tilde{I}_t(\bu) = \frac{1}{s_m}\sum_{j=1}^{s_m} & \frac{{\beta_m}\left(\bomega^{(j)}\right)  C(\bx, \bomega^{(j)}, \viewdir) }{p^{(m)}\left(\bomega^{(j)}\right)} \nonumber \\
    +\frac{1}{s_\ell}\sum_{j=s_m\!+\!1}^{s_m\!+\!s_\ell} & \frac{ {\beta_\ell}\left(\bomega^{(j)}\right)  C(\bx, \bomega^{(j)}, \viewdir)}{p^{(\ell)}\left(\bomega^{(j)}\right)}\,, \nonumber \\
    & \text{with } \bomega^{(1)}, \ldots , \bomega^{(s_m)} \sim p^{(m)}(\bomega)\,, \nonumber\\
    & \text{and } \bomega^{(s_m + 1)}, \ldots , \bomega^{(s_m + s_\ell)} \sim p^{(\ell)}(\bomega)\,,
\end{align}
where $\beta_m$ and $\beta_\ell$ are the power heuristic weights for multiple importance sampling, with exponent $2$, as in~\cite{veach1995optimally}:
\begin{align}
    \beta_m(\bomega) &= \frac{\left(s_m p^{(m)}(\bomega)\right)^2}{\left(s_m p^{(m)}(\bomega)\right)^2 + \left(s_\ell p^{(\ell)}(\bomega)\right)^2}, \\
    \beta_\ell(\bomega) &= \frac{\left(s_\ell p^{(\ell)}(\bomega)\right)^2}{\left(s_m p^{(m)}(\bomega)\right)^2 + \left(s_\ell p^{(\ell)}(\bomega)\right)^2}.
\end{align}
Similar to the observation of Zeltner \etal~\cite{zeltner2021MonteCarlo}, our experiments show that it is beneficial to ``detach'' gradients from the sampling procedure for $\bomega^{(j)}$ as well as from the PDFs $p^{(m)}$ and $p^{(\ell)}$ in Equations~\ref{eq:pl} and~\ref{eq:pm}.

In order to generate samples from the lighting and material distributions, we use inverse transform sampling, where the $n$ input pairs of samples $(u_0, v_0), \ldots, (u_{n-1}, v_{n-1}) \in [0, 1]^2$ are computed from $2n$ i.i.d.\ uniform random variables $r_0, \ldots, r_{n-1}, t_0, \ldots t_{n-1} \sim\text{Uniform}[0, 1]$ using:
\begin{align}
    u_i = \frac{\text{mod}(i, s) + r_i}{s}\\
    v_i = 2\cdot\frac{\lfloor i / s\rfloor + t_i}{s}\,,
\end{align}
where:
\begin{equation}
    s = 2^{\left\lfloor \frac{\log_2(n)+1}{2} \right\rfloor}.
\end{equation}
This procedure divides the unit square into an $\frac{s}{2}\times s$ grid, and uniformly samples a single pair $(u_i, v_i)$ in each one of the grid cells, resulting in a stratified sampling pattern.

For generating the data, we repeat the entire rendering process described above $16$ times using different primary and secondary rays, for every pixel and average the results in order to antialias the results. Note that this corresponds to using a box reconstruction filter, which may have visible artifacts, yet we decided to use it in our experiments for simplicity.

\def\tatertop{1.1in}
\begin{figure*}
\centering
\begin{tabular}{@{}c@{\,}c@{\,}c@{\,}c@{\,}c@{}}
Original illumination, $\alpha=0.8$ & Relit, $\alpha=0.6$ & Relit, $\alpha=0.4$ & Relit, $\alpha=0.2$ \\
\includegraphics[trim={0in \tatertop{} 0in \tatertop{}}, clip, width=0.24\linewidth]{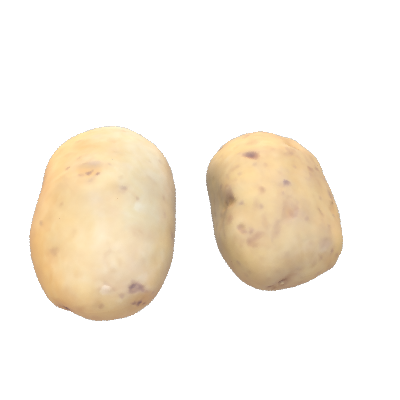} &
\includegraphics[trim={0in \tatertop{} 0in \tatertop{}}, clip, width=0.24\linewidth]{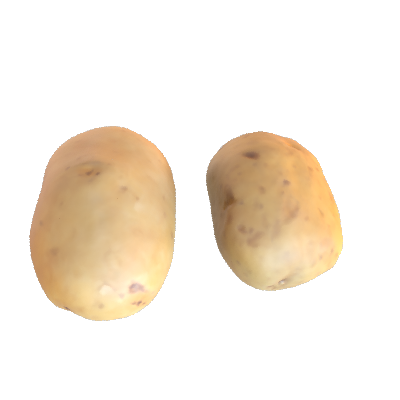} &
\includegraphics[trim={0in \tatertop{} 0in \tatertop{}}, clip, width=0.24\linewidth]{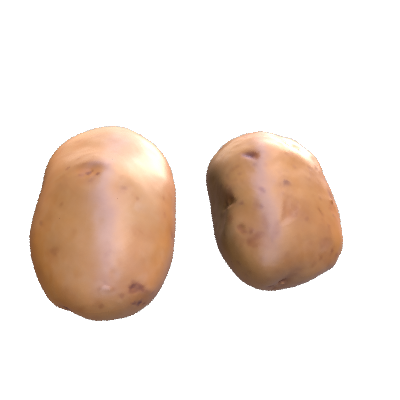} &
\includegraphics[trim={0in \tatertop{} 0in \tatertop{}}, clip, width=0.24\linewidth]{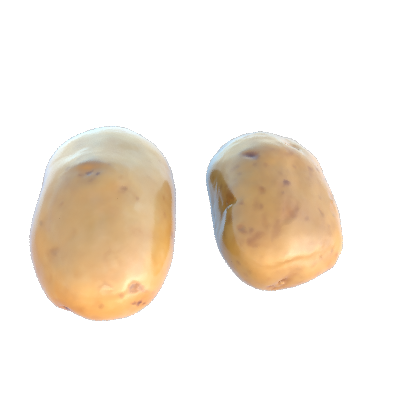}
\end{tabular}
\caption{Our method enables modifying the illumination and material properties of the object. On the left, we show the original illumination and rough material ($\alpha=0.8$ in our BRDF model). The three columns to the right feature our recovered albedo, but rendered under different environment maps and progressively lower roughness values.}
\label{fig:relitpotato}    
\end{figure*}

\begin{figure*}
\centering
\begin{tabular}{@{}c@{\,}c@{\,}c@{\,}c@{\,}c@{}}
Sample image & Illumination & Albedo $\brho$ & Roughness $\alpha$ & Fresnel $\kappa$ \\
\rotatebox{90}{\quad\quad Recovered}
\includegraphics[width=0.16\linewidth]{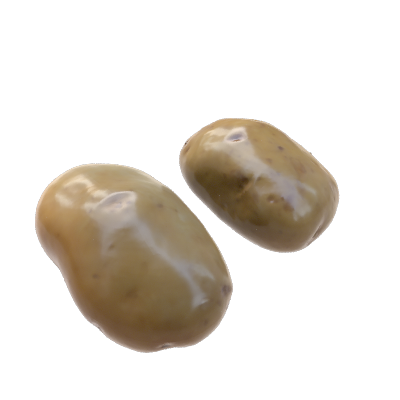} &
\includegraphics[width=0.32\linewidth]{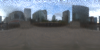} &
\includegraphics[width=0.16\linewidth]{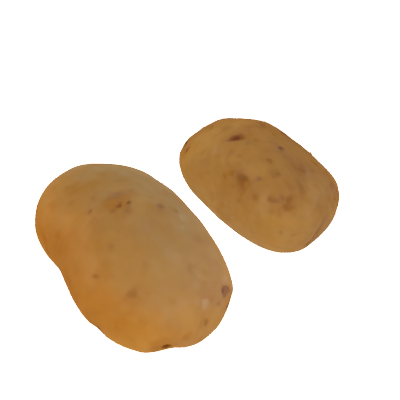} &
\includegraphics[width=0.16\linewidth]{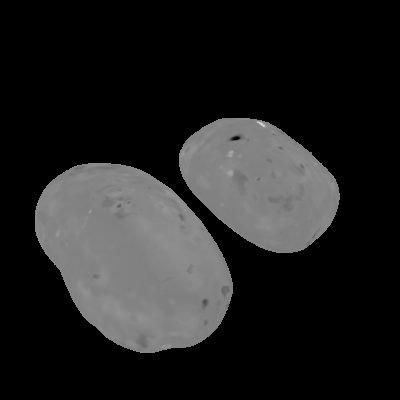} &
\includegraphics[width=0.16\linewidth]{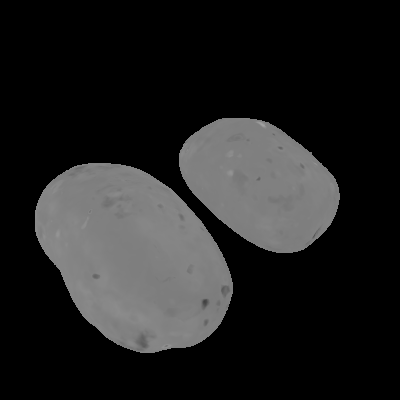} \\
\rotatebox{90}{\quad Ground truth}
\includegraphics[width=0.16\linewidth]{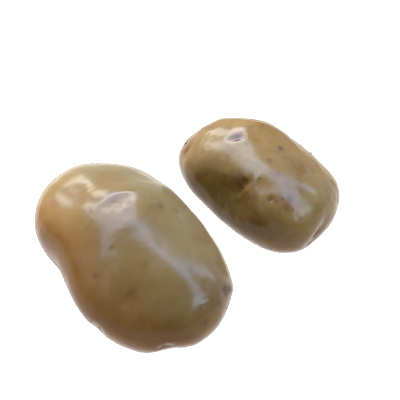} &
\includegraphics[width=0.32\linewidth]{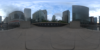} &
\includegraphics[width=0.16\linewidth]{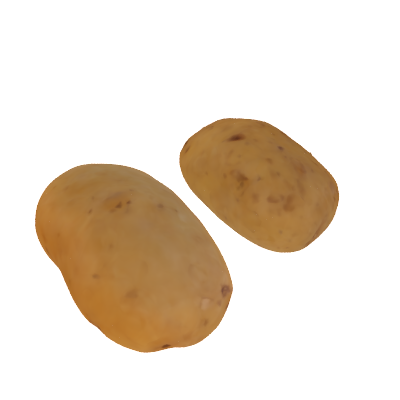} &
\includegraphics[width=0.16\linewidth]{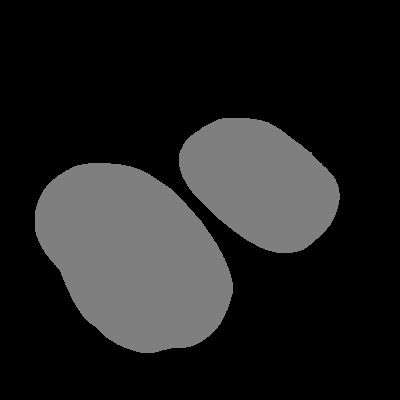} &
\includegraphics[width=0.16\linewidth]{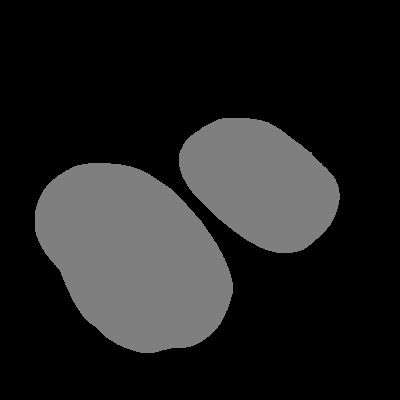}
\end{tabular}
\caption{Our results on a shinier object. Our method is designed for diffuse objects, but it still manages to extract accurate material properties and illumination when materials are more specular ($\alpha = 0.2$ for this figure). For visualization purposes the roughness values $\alpha$ are linearly mapped from  $[0.15, 0.25]$ to $[0, 1]$, and the specular reflectance at normal incidence $\kappa$ are mapped from $[0.03, 0.05]$ to $[0, 1]$ (with the true value being $0.04$), in order to show small errors.}
\label{fig:shinypotato}    
\end{figure*}

\begin{figure}
    \centering
    \begin{tabular}{@{}c@{\,}c@{}}
    \includegraphics[width=0.49\linewidth]{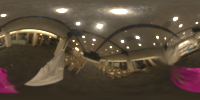} &
    \includegraphics[width=0.49\linewidth]{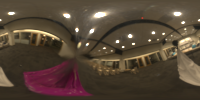}
    \end{tabular} 
    \footnotesize (a) Sample environment maps from the same captured scene \\
    \vspace{0.1in}
    \begin{tabular}{@{}c@{\,}c@{}}
    \includegraphics[width=0.49\linewidth]{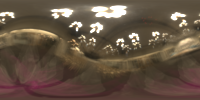} &
    \includegraphics[width=0.49\linewidth]{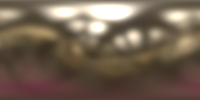} \\
    \footnotesize (b) Mean environment map & (c) Blurred mean environment map 
    \end{tabular} 
    \vspace{-0.1in}
    \caption{Our blurring process for environment maps provided with the Stanford-ORB dataset~\cite{kuang2023stanford}. The environment maps provided with the dataset are not aligned since they were captured by a light probe not co-located with the object, as shown by the two misaligned images in panel (a), which makes them unusable for computing error metrics for illumination recovery. We therefore first: (b) average all environment maps provided along with each scene, and then (c) blur them using a normalized spherical Gaussian.
    \label{fig:envmapblur}
    }
\end{figure}

\section{Theoretical Considerations}

\subsection{Figure~3 from Main Paper}

Figure~3 from the main paper is based on a simplified ``flatland'' version of our problem, where the occluders and materials are known. A circular object with radius $1$ and Lambertian BRDF with albedo $1$ is placed at the origin, and illuminated by a (monochromatic) environment map $L(\phi)$. An occluder of some angular width $\Delta\theta$ and central angle $\theta$ is then placed at some radius $r$ from the origin, blocking all shadow rays from the object to the illumination.

We parameterize the surface of the object using a single angle parameter $\lambda \in [0, 2\pi)$, \ie the ``surface'' of the disk is the set of points $\mathcal{S} = \{(\cos\lambda, \sin\lambda): \lambda \in [0, 2\pi)\}$. The ``image'' at the point $\lambda$ at time $t$ is then:
\begin{equation} \label{eq:flatrender}
    I_t(\lambda) = \int_{0}^{2\pi} L(\phi) M_t(\lambda, \phi; \theta_t, \Delta\theta_t)(\cos(\lambda-\phi))_+ \, d\phi\,,
\end{equation}
with the ``flatland'' (unnormalized) BRDF set to $1$, and the occluder value at time $t$ is:
\begin{align}
&M_t(\lambda, \phi; \theta_t, \Delta\theta_t) = \mathbbm{1}\!\left[\left|\psi(\lambda, \phi) - \theta_t \right| < \frac{\Delta\theta_t}{2}\right],
\end{align}
with:
\begin{align}
&\psi(\lambda, \phi) = \arctan\!\left(\frac{\sin\lambda + t(\lambda - \phi)\sin\phi}{\cos\lambda + t(\lambda - \phi)\cos\phi}\right), \\
&t(\delta) = \sqrt{\cos^2(\delta)+r^2-1} - \cos(\delta). \nonumber
\end{align}

We then discretize the integral in Equation~\ref{eq:flatrender} for every $t=1, \ldots, T$, and stack the resulting equations:
\begin{equation} \label{eq:matrixell}
    \underbrace{\begin{bmatrix}\mathbf{I}_1\\ \vdots \\\mathbf{I}_T\end{bmatrix}}_{\mathbf{I}} = \underbrace{\begin{bmatrix}A_1\\ \vdots \\A_T\end{bmatrix}}_{A}\boldsymbol{\ell}, 
\end{equation}
where $\mathbf{I}_t\in\R^N$ is a vector of rendered values $\{I_t(\lambda)\}$ at time $t$, $\boldsymbol{\ell}\in\R^M$ is a vector of illumination values $\{L(\phi)\}$, and $A_t\in\R^{N\times M}$ contains the corresponding values of the occluders $M(\lambda, \phi; \theta_t, \Delta\theta_t)$ at time $t$ and the cosine lobe $(\cos(\lambda-\phi))_+$.

Figure~3 then shows the singular values of $A$ (which are the square roots of the eigenvalues of $A^\top A$), normalized to have maximum value $1$, for a few scenarios:
\begin{itemize}
    \item ``No occlusions'': The scenario described above with a single observation ($T=1$) but without an occluder, \ie where the mask is set to $1$ everywhere (or, alternatively, $\Delta\theta = 2\pi$).
    \item ``1 observation'': The scenario described above, with a single occluder ($T=1$) placed at $\theta=0$.
    \item ``$T$ observations'', for $T\in\{2, 8, 16, 32\}$: $A$ contains $T$ stacked matrices corresponding to a discretization of Equation~\ref{eq:flatrender}, each one for a different occluder location $\theta_t$ uniformly spread over $[0, 2\pi)$, and the same $\Delta\theta$.
\end{itemize}

The exact values used for Figure~3 are occluder width $\Delta\theta = 0.7 \text{ rads}$, placed at distance $r=10$ from the origin, with $512$ samples per observation.

\subsection{Spherical Harmonics Intuition}

In the main paper we show that expressing the occluder masks in the basis of spherical harmonics, instead of the standard basis, leads to significantly improved results. The reasoning behind that can be illustrated in 1D using the Fourier basis (sines and cosines), since spherical harmonics are the spherical equivalent of the Fourier basis.

Similar to Equation~\ref{eq:matrixell}, the integral in Equation~\ref{eq:flatrender} can also be discretized and combined into a linear equation in the occluder values. However, the resulting matrix is block-diagonal, with each $M_t$ only depending on $\mathbf{I}_t$ at the same time $t$:
\begin{equation} \label{eq:matrixem}
    \mathbf{I}_t = B\mathbf{m}_t,
\end{equation}
where $\mathbf{I}_t$ is again a vector of image values at time $t$, $\mathbf{m}_t$ is a vector of occluder values at the same time, and $B$ describes their linear relation according to a discretization of Equation~\ref{eq:flatrender}. $B$ is composed of a circulant matrix (corresponding to the BRDF's convolution), multiplied by a diagonal matrix with the illumination values $L$ along its diagonal. This means that when the illumination is uniform, $B$ is also a circulant matrix, and is therefore diagonalized by the Discrete Fourier Transform (DFT) matrix. This makes the problem diagonal in the Fourier basis, which makes adaptive optimizers such as Adam~\cite{kingma2015adam} especially effective.

\begin{figure}
    \centering
    \begin{tabular}{@{}c@{}c@{}}
    & Light spectrum \quad\quad\quad Row 32 \quad\quad\quad\quad Row 64 \quad\quad \\[-0.4ex]
    \rotatebox{90}{\;\quad\quad $a=10$ \quad\quad\quad\, $a=1$ \quad\quad\quad $a=0.1$} & \includegraphics[width=\linewidth]{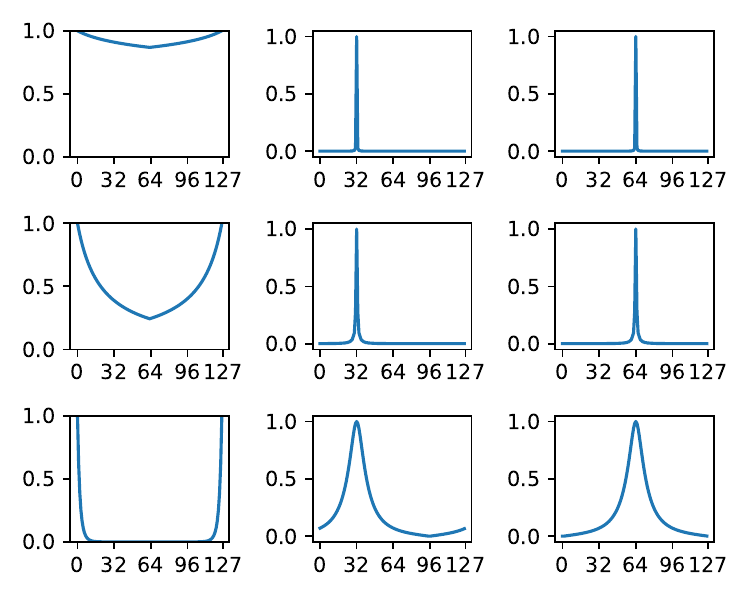}
    \end{tabular}
    \caption{Most of the energy of the matrix $B$ is concentrated around the diagonal elements in the Fourier basis. Each row shows, from left to right: the incident illumination spectrum (which modifies $B$), and the magnitude of the elements of the $32$nd and $64$th rows of $FBT^\top$, where $F$ is the DFT matrix. Each row corresponds to a $1/f^a$ spectrum for $a = 0.1, 1, 10$. See text for more details.}
    \label{fig:almostdiagonal}    
\end{figure}

When the illumination is non-uniform, $B$ is not generally diagonalized by the DFT matrix. However, for natural lighting, when the DFT matrix $F$ is applied to $B$, most of the matrix's energy is along the diagonal elements, \ie the elements of the $i$th row of $FBF^\top$ are maximized by the $i$th element. See Figure~\ref{fig:almostdiagonal} for examples showing rows of the $FBF^\top$ matrix for different illumination spectra. The fact that the Fourier basis (or in the original problem, the basis of spherical harmonics) ``nearly-diagonalizes'' the problem is the reason for its effectiveness.

Furthermore, since we are not interested in recovering the occluders in regions with low illumination (see right of Figure~5 of the main paper), it is informative to consider the problem of estimating the masked illuminant:
\begin{equation} \label{eq:matrixellem}
    \mathbf{I}_t = C(\boldsymbol{\ell}\circ\mathbf{m}_t),
\end{equation}
where $\circ$ denotes elementwise multiplication, and $C$ is a circulant matrix corresponding to the (discretized) convolution in Equation~\ref{eq:flatrender}. The problem of estimating $\boldsymbol{\ell}\circ\mathbf{m}_t$ given $\mathbf{I}_t$ is in fact diagonalized by the Fourier basis, because $C$ is circulant.

\subsection{Image Pyramid Intuition}

\begin{figure}
\centering
\begin{tabular}{@{}c@{\,}c@{\,}c@{}}
$T=1$ & $T=2$ & $T=4$ \\
\includegraphics[width=0.32\columnwidth]{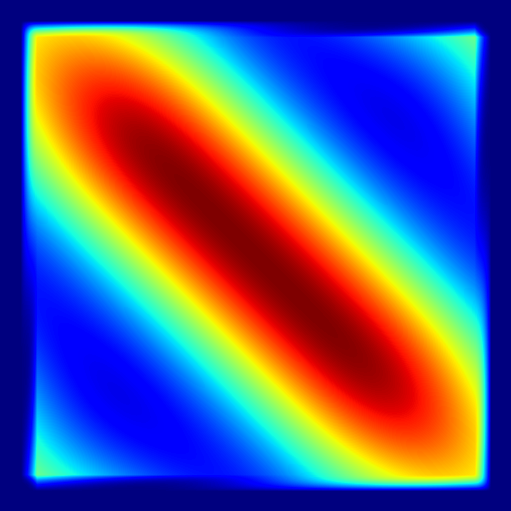} &
\includegraphics[width=0.32\columnwidth]{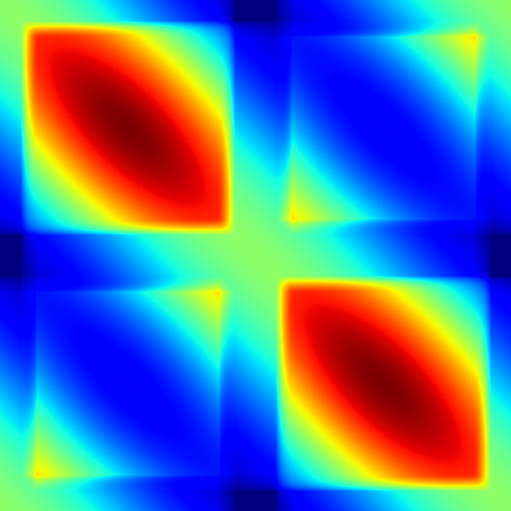} &
\includegraphics[width=0.32\columnwidth]{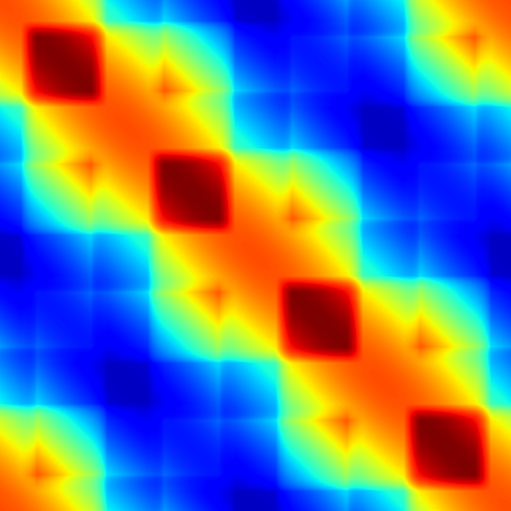} \\
$T=8$ & $T=16$ & $T=32$ \\
\includegraphics[width=0.32\columnwidth]{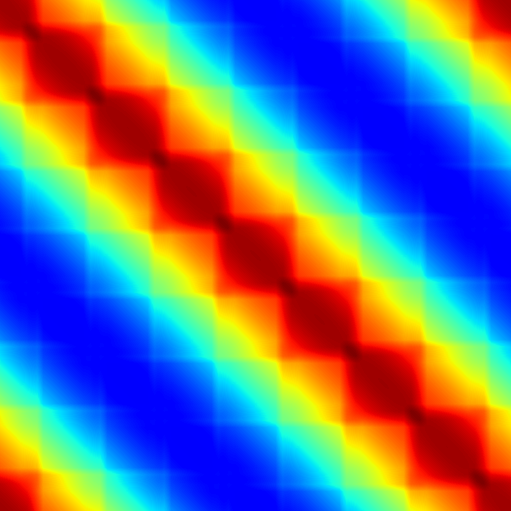} &
\includegraphics[width=0.32\columnwidth]
{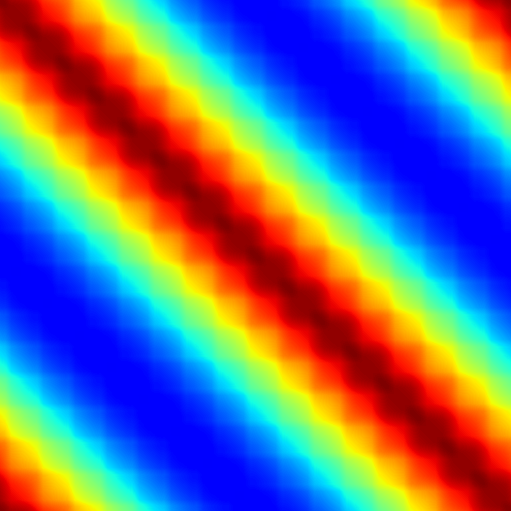} &
\includegraphics[width=0.32\columnwidth]{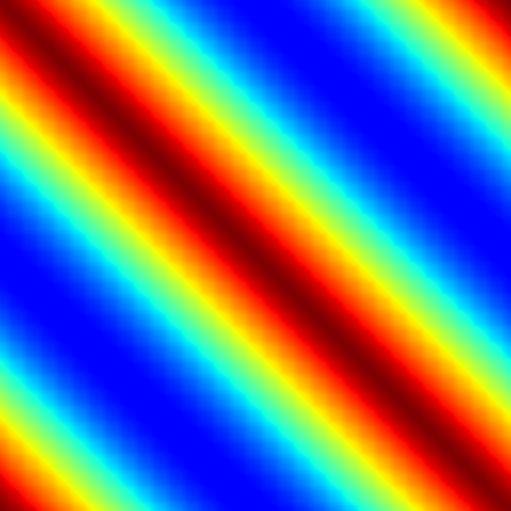}
\end{tabular}
\caption{The structure of $A^\top A$ for the light transport matrix $A$, plotted for different values of the number of observations $T$. Large values in red, small values in blue.}
\label{fig:ATA}    
\end{figure}

The set of linear equations satisfied by the illumination $\boldsymbol{\ell}$ described in Equation~\ref{eq:matrixell} cannot be separately solved for each $t$, and therefore we must consider the entire system of equations described by the matrix $A\in\R^{NT\times M}$. Figure~\ref{fig:ATA} shows the entries of $A^\top A$, for different choices of $T$. The entries of the matrix exhibit a block structure in the standard basis, \ie without applying the DFT matrix to it, with the sizes of the blocks becoming smaller for larger values of $T$. In fact, for very large values of $T$ the matrix approaches a circulant matrix, which is diagonalized by the DFT matrix. However, we wish our method to not make use of that fact and work even for a sparse set of observations, and therefore a natural choice for parameterizing the illumination $\boldsymbol{\ell}$ is by using an image pyramid, which matches the block structure of $A^\top A$ for general $T$ values.

\section{Geometry Estimation Method}

We estimate unknown object geometry from input images by recovering a Neural Radiance Field (NeRF)~\cite{mildenhall2020nerf} of the object, based on the Instant Neural Graphics Primitives~\cite{mueller2022instant} representation. To improve the recovered normal vectors (computed as the negative normalized gradient of the volume density field), we use the normals orientation regularization and MLP-predicted normals technique from Ref-NeRF~\cite{verbin2022refnerf}.

\section{Data Specification}

All BRDF parameters (RGB albedo, roughness, and the specular reflectance at normal incidence), are output by a coordinate-based MLP taking in a positionally-encoded location:
\begin{align}
    \gamma(\bx) = \big(&\sin(\bx),\cos(\bx), \sin(2\bx), \cos(2\bx), \nonumber\\
    &\ldots, \sin(64\bx), \cos(64\bx)\big).
\end{align}

The MLP has $4$ layers with $128$ hidden units each, with ReLU nonlinearities. The weights are initialized around zero, and the output BRDF parameters are obtained by mapping the MLP's output through a sigmoid function, meaning that they are all initialized around a value of $0.5$.

The spherical harmonic coefficients $\{a_{t\ell m}\}$ from Equation~6 of the main paper are all set to zero, but a positive bias of $100$ is added to the pre-sigmoid value (which can equivalently be done by initializing $a_{t00}$ to a constant). We find that initializing the masks to be close to $1$ everywhere improves our method's performance and prevents it from getting stuck in a local minimum.

The environment map pyramid levels are also all initialized to zero, such that the illuminant is set to all-ones, due to the exponential nonlinearity (see Equation~7 in the main paper).

\section{Additional Data Details}

The geometry of the objects in the paper and their textures originated in the following BlendSwap models:
\begin{enumerate}
    \item Potatoes: created by \emph{mik1190}, CC0 license (model \#15725)
    \item Chair: created by \emph{1DInc}, CC0 license (model \#8261).
    \item Mannequin: created by \emph{salimrached}, CC0 license (model \#27747).
    \item Toad: created by \emph{arenyart}, CC0 license (model \#13078).
    \item Plant: created by \emph{New Enemy}, CC0 license (model \#30071).
    \item Giraffe: created by \emph{amx360}, CC-BY license (model \#29651).
\end{enumerate}

The environment maps are from the following Poly Haven assets:
\begin{enumerate}
    \item ``Canary Wharf'': created by \emph{Andreas Mischok}, CC0 license.
    \item ``Abandoned Factory Canteen 01'': created by \emph{Sergej Majboroda}, CC0 license.
    \item ``Outdoor Umbrellas'': created by \emph{Sergej Majboroda}, CC0 license.
    \item ``Thatch Chapel'': created by \emph{Dimitrios Savva, Jarod Guest}, CC0 license.
    \item ``Evening Road 01 (Pure Sky)'': created by \emph{Jarod Guest, Sergej Majboroda}, CC0 license.
    \item ``Marry hall'': created by \emph{Sergej Majboroda}, CC0 license.
\end{enumerate}

\end{document}